\documentclass[letterpaper]{article} 
\usepackage{aaai24}  
\usepackage{times}  
\usepackage{helvet}  
\usepackage{courier}  
\usepackage[hyphens]{url}  
\usepackage{graphicx} 
\usepackage{natbib}  
\usepackage{caption} 
\usepackage[ruled]{algorithm2e}
\usepackage{subfigure}
\usepackage{tabularx}
\usepackage{multirow}
\usepackage{mathtools}
\usepackage{xspace}
\usepackage{enumitem}
\usepackage{soul}

\usepackage{appendix}

\title{A General Implicit Framework for Fast NeRF Composition and Rendering}
\author{
    Xinyu Gao\textsuperscript{\rm 1},
    Ziyi Yang\textsuperscript{\rm 1},
    Yunlu Zhao\textsuperscript{\rm 1},
    Yuxiang Sun\textsuperscript{\rm 2},
    Xiaogang Jin\textsuperscript{\rm 1}\thanks{Corresponding author.},
    Changqing Zou\textsuperscript{\rm 1,2}
}
\affiliations{
    \textsuperscript{\rm 1}State Key Lab of CAD\&CG, Zhejiang University,
    \textsuperscript{\rm 2}Zhejiang Lab \\
    
    \{22121052, Ingram14, yunlu.zhao\}@zju.edu.cn, sunyuxiangyx@gmail.com \\
    jin@cad.zju.edu.cn, changqing.zou@zju.edu.cn \\
}

\begin{document}

\maketitle

\begin{abstract}
A variety of Neural Radiance Fields (NeRF) methods have recently achieved remarkable success in high render speed.
However, current accelerating methods are specialized and incompatible with various implicit methods, preventing real-time composition over various types of NeRF works. Because NeRF relies on sampling along rays, it is possible to provide general guidance for acceleration. To that end, we propose a general implicit pipeline for composing NeRF objects quickly.
Our method enables the casting of dynamic shadows within or between objects using analytical light sources while allowing multiple NeRF objects to be seamlessly placed and rendered together with any arbitrary rigid transformations. Mainly, our work introduces a new surface representation known as Neural Depth Fields (NeDF) that quickly determines the spatial relationship between objects by allowing direct intersection computation between rays and implicit surfaces. It leverages an intersection neural network to query NeRF for acceleration instead of depending on an explicit spatial structure.Our proposed method is the first to enable both the progressive and interactive composition of NeRF objects. Additionally, it also serves as a previewing plugin for a range of existing NeRF works.
\end{abstract}

\begin{figure*}[htbp]
    \centering
    \begin{minipage}{\linewidth}
      \includegraphics[width=0.45\linewidth]{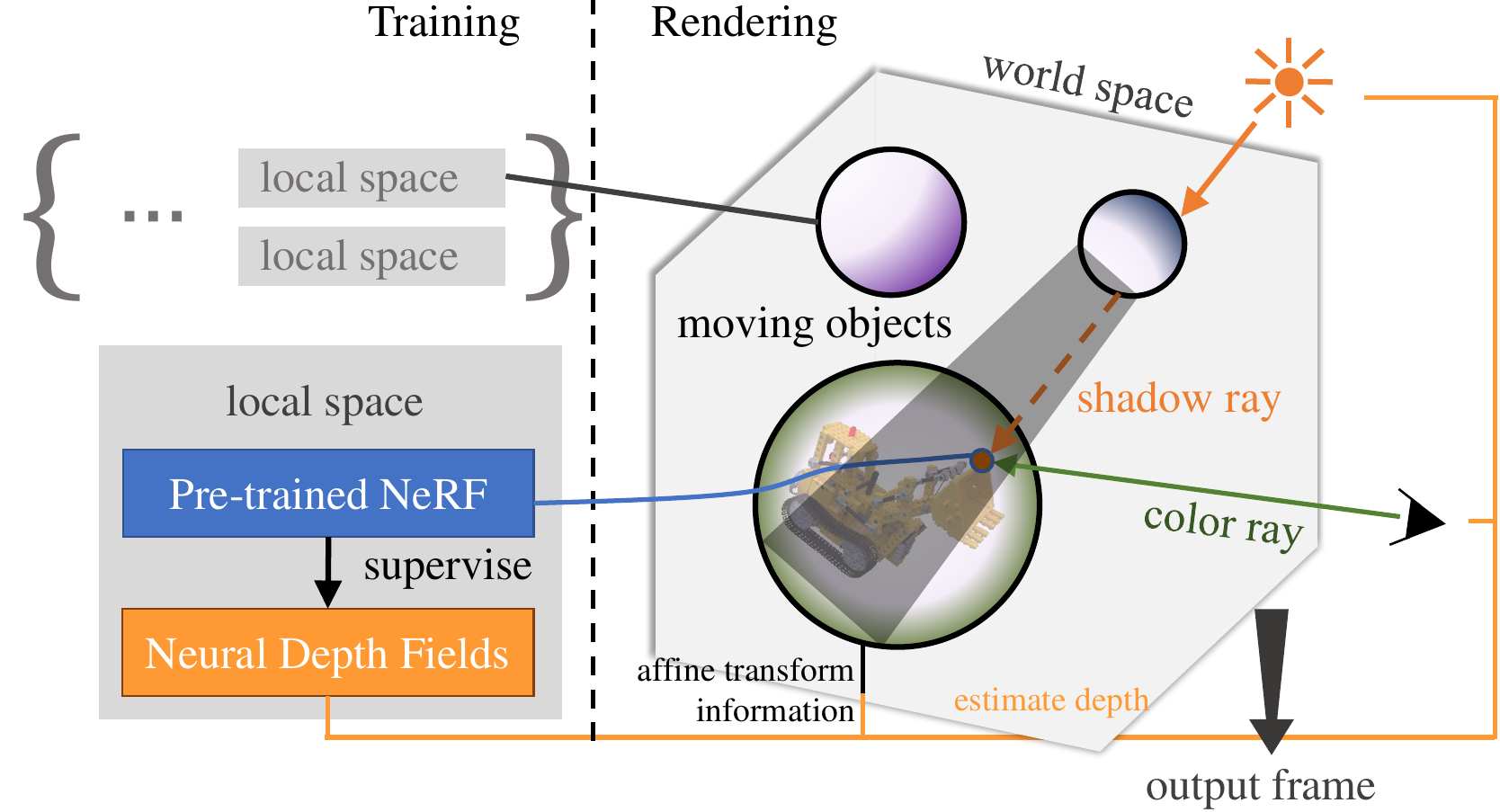}
      \includegraphics[width=0.55\linewidth]{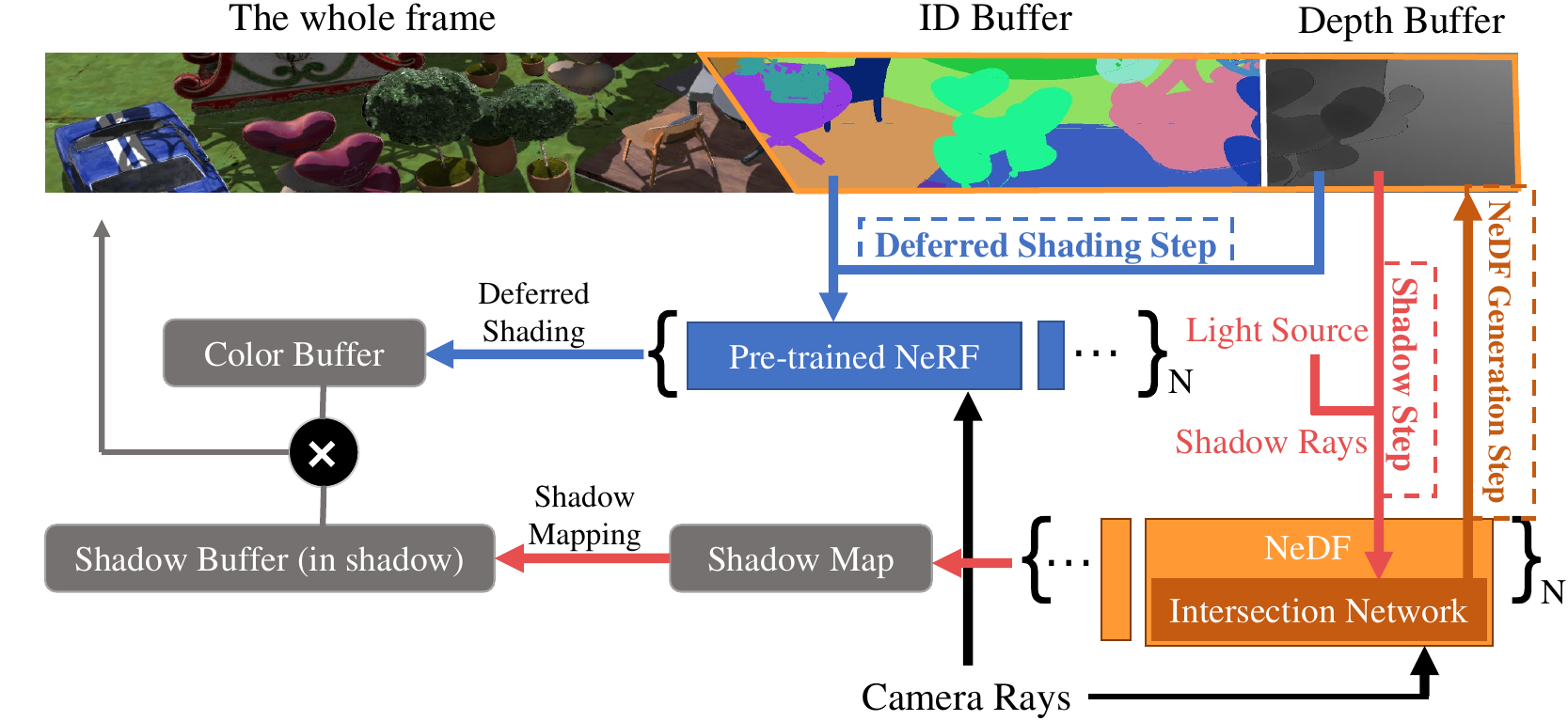}
    \end{minipage}
    \caption{An overview of our framework (left) and data flow (right). In terms of framework, we use Neural Depth Fields (NeDF) as the representation of the implicit surface within each object's local space during training, under the supervision of pre-trained NeRF. There are three main steps in render time: 1) Step1 is "NeDFs generation step", depth buffer and ID buffer are created by a collection of NeDFs; 2) Step2 is "Deferred shading step", those 2 buffers are then used by a collection of NeRFs to fill a color buffer with no shadows; 3) Step3 is "Shadow step", shadow rays are generated based on the depth buffer and NeDFs are used again to provide a shadow map. Finally, the color buffer and shadow buffer are multiplied to produce the results.}
    \label{fig:overview}
\end{figure*}

\section{Introduction}
The development of virtual worlds has revealed a strong preference for neural techniques.
Neural Radiance Fields (NeRF), an image-based rendering method~\cite{mildenhall2020nerf}, has shown its proficiency in rendering with high fidelity while also offering the added advantage of compression capabilities, resulting in a low storage footprint. As a result, NeRF has been extensively studied in human face editing \cite{wang2022morf, jiang2022nerffaceediting}, autonomous driving \cite{tancik2022block, kundu2022panoptic}, stylization \cite{gu2021stylenerf}, mesh reconstruction \cite{wang2021neus, sun2022neural}, inverse rendering \cite{boss2021nerd, boss2021neural, zhang2021nerfactor}, and so on.

Although a number of NeRF composition methods \cite{Ma_Li_He_Dong_Tan_2023, yang2021learning, wu2022object, shuai2022multinb, ost2021neural} have achieved impressive results and improved the flexibility by modeling a scene as multiple semantic parts rather than as whole, compositing neural radiance fields in real-time still remains under-explored.
Existing acceleration methods, such as implicit methods \cite{kurz2022adanerf, neff2021donerf} that rely on a limited scope of camera activity, hybrid methods \cite{liu2020neural, mueller2022instant} that require a complex spatial structure for querying, or pre-computing methods \cite{hedman2021baking, chen2022mobilenerf} that incur significant storage costs, are far from straightforward in this task due to various challenges, let alone forming a general pipeline.
First, it is difficult to satisfy low memory-storage cost and real-time speed while maintaining arbitrary affine transformations for implicit objects. 
Second, as the number of objects to be composed grows (e.g., a simple duplicate of 1k balls), rendering the entire scene (especially considering shadows among implicit objects) takes long, and the cost of hybrid or pre-computing methods becomes prohibitively expensive. 

To address the aforementioned issues, we present a general implicit framework that can quickly perform compositing and shadow casting of NeRF objects without querying any complex structure, thereby avoiding the excessive memory-storage costs. Meanwhile, we propose Neural Depth Fields (NeDF), a novel implicit representation for implicit surface that aims for fast and precise depth estimation under a flexible camera activity scope and arbitrary affine transformations of objects, overcoming critical issues associated with previous implicit acceleration methods while retaining the benefits of implicit methods with a low storage cost. When working with a single object in a composed scene, manipulation speeds can even reach real-time in our CUDA implementation. Notably, we use the classic concept of shadow mapping based on NeDF for efficient rendering of dynamic shadows cast by either point light or parallel light sources, which represents a significant advancement over previous NeRF works aimed at acceleration.
\par
Our work is able to provide users a fast NeRF scene composition tool allowing a quick preview of the composited results, as users may prefer interactive-level speed over perfect quality when manipulating each NeRF object. The key technical contributions are summarized below:
\begin{itemize}
    \item A novel framework that enables interactive previewing of the NeRF composition process at a low storage cost, thereby providing a new interactive way to create novel realistic scenes from real-world scenes.
    \item A quick and efficient method for generating intra and inter-object dynamic shadows cast by either point light or parallel light sources.
    \item A new geometry representation for NeRF objects that aids intersecting/depth estimation during arbitrary manipulation.
\end{itemize}

\section{Related Work}
\paragraph{Compositing Neural Radiance Fields.}
Modeling a scene as a single NeRF remains generally limited in functionality. In many scenarios, the ability to isolate individual objects and manipulate them under controlled constraints is necessary (e.g., driving scenes \cite{krishnan2023lane}). In order to enhance the controllability and editability of the NeRF-based virtual scenes, one straightforward idea is to model scenes as multiple semantic parts rather than a whole at the beginning. This can be achieved by separately compressing background and foreground objects into two branches represented by neural voxels \cite{yang2021learning} or a set of neural SDFs \cite{wu2022object}. In order to enable a better perception for foreground objects, clues such as 3D motion \cite{shuai2022multinb} or segmentation mask \cite{kundu2022panoptic, tancik2022block} in 2D image space may be required as a preprocessing step. Another idea is compositing pre-trained NeRF models together to form a more complex and controllable scene. To achieve this, explicit spatial structures like multi-granularity voxels \cite{liu2020neural} and 3D bounding box \cite{ost2021neural, shuai2022multinb} are employed to indicate the sampling range for each nerf model in global space. By manipulating these voxels/boxes, NeRF objects are transformed from their local space coordinates to the global world space. Additionally, techniques that pre-compute radiance in explicit structures have shown promise for scene compositing. Despite the benefits of utilizing multiple neural fields, doing so may result in significant computational requirements, leading to notable storage overhead in the pre-compute methods or reduced rendering speeds in implicit methods, which can negatively affect the ease with which users can manipulate objects in an interactive preview and loading speed. This limitation contradicts the goal of enabling real-time scene editing and rendering. Unlike these methods, we attempt to composite NeRF objects interactively and progressively for a quick preview.

\paragraph{Accelerating Rendering of NeRF-modeled Scenes}
Real-time rendering has been a long-standing goal in the field of computer graphics, and this objective extends to the realm of NeRF. The explicit (pre-computed) methods achieved real-time by pre-computing radiance color as spherical Harmonics coefficients \cite{yu2021plenoctrees} or feature vectors \cite{hedman2021baking,chen2022mobilenerf}, and then storing them in spatial structures like octree for quick inference, thereby reducing the computational overhead at render time. Among these explicit methods, the state-of-the-art method \cite{bakedsdf2023sig} converts NeRF scenes into a combination of meshes and spherical Gaussians, resulting in fast rendering speed on modern mobile graphics pipelines. However, the explicit method can result in high storage costs for even a single scene. In contrast, implicit (MLP-based) methods that mainly focus on importance sampling \cite{piala2021terminerf, neff2021donerf, kurz2022adanerf, Niemeyer2021Regnerf} significantly reduce storage overhead and computational burden. But these methods can limit the activity scope of the camera to a specific view cell. The hybrid method \cite{liu2020neural} incorporates a neural component within the explicit grid to achieve a cost-effective solution. All of these methods are highly specialized, designed to address their own accelerating path. Our goal, on the other hand, is to investigate a general pipeline that enables rapid composition across various NeRF works.

\paragraph{Neural Light Fields (NeLF)}
The representation of Light Fields has been widely utilized in computer graphics to model a scene as a 4D representation, enabling fast image-based rendering. The emergence of deep learning in recent years has led to the investigation of Neural Light Fields (NeLF) as an extension of this concept
\cite{meng2019high,wizadwongsa2021nex,attal2022learning,sitzmann2021light,suhail2022light}. Recently, R2L \cite{cao2022real, wang2022r2l} successfully obtained NeLF with a comparable quality to NeRF, which provides a straightforward insight into the capability of light fields' representation to generate high-quality results under $360^{\circ}$ cases. In comparison to NeRF, NeLF has the distinct advantage of requiring only a single evaluation per ray, thereby eliminating the need for time-consuming ray marching. However, the composition task is unsuitable for NeLF due to its lack of precise depth information, though several past approaches \cite{1328805, 6910019, Kim2013scene} have tried to leverage light fields to extract geometry information for depth estimation. Nevertheless, NeLF has inspired us to efficiently composite NeRF in a geometry-aware manner.

\section{Methodology}

Given a set of objects represented by NeRF, our approach provides a quick preview of the resulting scene in which all of the NeRF objects are manipulated with arbitrary affine transformations and rendered together. The overview and data flow of our render framework is shown in Fig. \ref{fig:overview}, which is also described by the  algorithm pseudo-codes in appendix. 

\begin{figure}[tbp]
  \centering
  \subfigure[]{
    \includegraphics[width=0.35\linewidth]{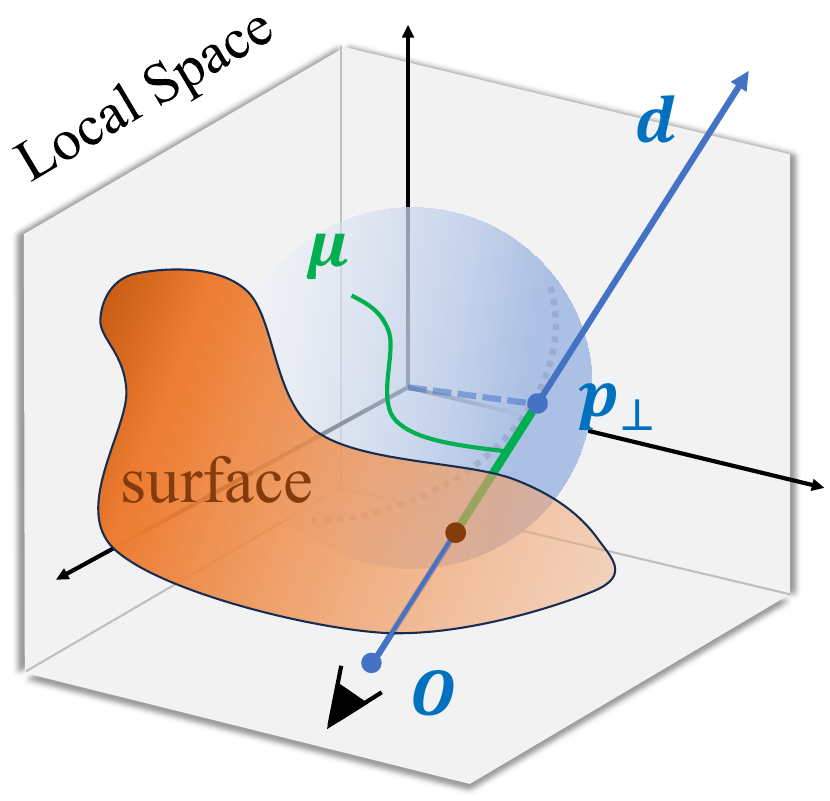}
  }
  \subfigure[]{
    \includegraphics[width=0.55\linewidth]{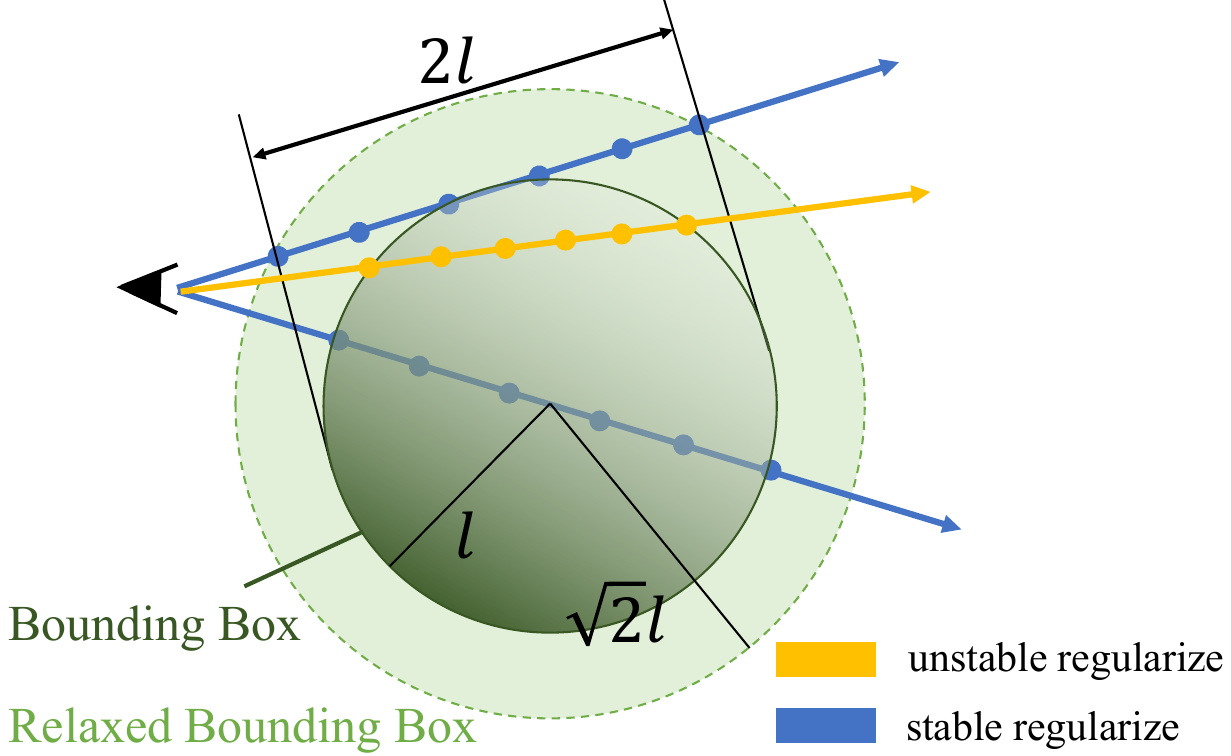}
  }
  \caption{(a) Given a ray traveling through an object's local space, the first intersection on the surface can be obtained by the distance between the intersection and the tangency point $\mathbf{p}_{\perp}$ (The sphere is in the center of local space).
  Since $\mathbf{p}_{\perp}$ can be quickly and uniquely determined, this representation leads to a direct intersection computation. (b) To improve the stability of NeDF, we regularize the main body of each ray, approximated by a tuple of points sampled along the ray, within the relaxed bounding box 
  before fedding it into the intersection network.
  }
  \label{fig:rep_ray}
\end{figure}

\subsection{Neural Depth Fields (NeDF)}
 Neural Depth Field is defined on a 5-dimensional continuous function $\mathcal{F}_{\Phi}$ that represents a scene. This function maps an arbitrary ray (represented by a starting point and a direction) to the distance/depth between the starting point of the ray $\mathbf{r}$ to the ray-object intersection point in a 3D virtual world space as shown in Fig. \ref{fig:rep_ray}a. Assuming that each object used for scene composition is opaque, this output distance can enable the computation of the surface point color of composed NeRF, as well as the relative spatial occlusion relationship of each composed object, which are essential for scene composition and rendering.

The continuous function $\mathcal{F}_{\Phi}$ is implemented with a Multilayer Perceptron network, termed as \textit{{intersection network.}} 
$\mathcal{F}_{\Phi}$ does not directly infer the depth between the starting point of the ray $\mathbf{r}$ and the ray-object intersection point. It instead infers the distance between the ray-object intersection point and the tangency point at which the ray $\mathbf{r}$ is tangent to a sphere centered at the origin in the local space of the object (i.e., {$\mathcal{F}_{\Phi}$} is used to compute $\mu$, and $\mu$ is the distance between the tangency point $\mathbf{p}_{\perp}$ of the ray $\mathbf{r}$ to the ray-object intersection point, as shown in Fig. \ref{fig:rep_ray}a).
We choose to learn $\mu$ rather than $D(\mathbf{r})$ is because it's impossible to learn an infinite space. $\mu$ is defined in the local space and has a limited range of values.

Specifically, for a given ray $\mathbf{r}$, we mathematically define $\mathcal{F}_{\Phi}$ as:
\begin{equation}
\mathcal{F}_{\Phi}:\mathbf{r} \mapsto (\mu,\hat{\alpha}),
\label{eq:NeDF_reg}
\end{equation}
where $\hat{\alpha}$ is a binary indicator that signifies whether the ray intersects the object's surface.

Once the $\mu$ is predicted by the intersection network, the depth $D(\mathbf{r})$ between the starting point of the ray $\mathbf{r}$ and the ray-object intersection point can be calculated as:
\begin{equation}
D(\mathbf{r})=|\mathbf{o}-\mathbf{p}_{\perp}|- \mu,
\label{eq:convert_depth}
\end{equation}
where $\mathbf{o}$ is the starting point of the ray and $\mathbf{p}_{\perp}$ is the tangency point. We adopt a set of points sampled along rays to represent an oriented ray to improve the stable performance of the intersection network. 
Specifically, we regularize the sampling within a specific local range along rays and sample 16 points in total (see Fig. \ref{fig:rep_ray}b). 

\subsection{Intersection network}
\begin{figure}[tbp]
  \centering
  \subfigure[]{
    \includegraphics[width=0.35\linewidth]{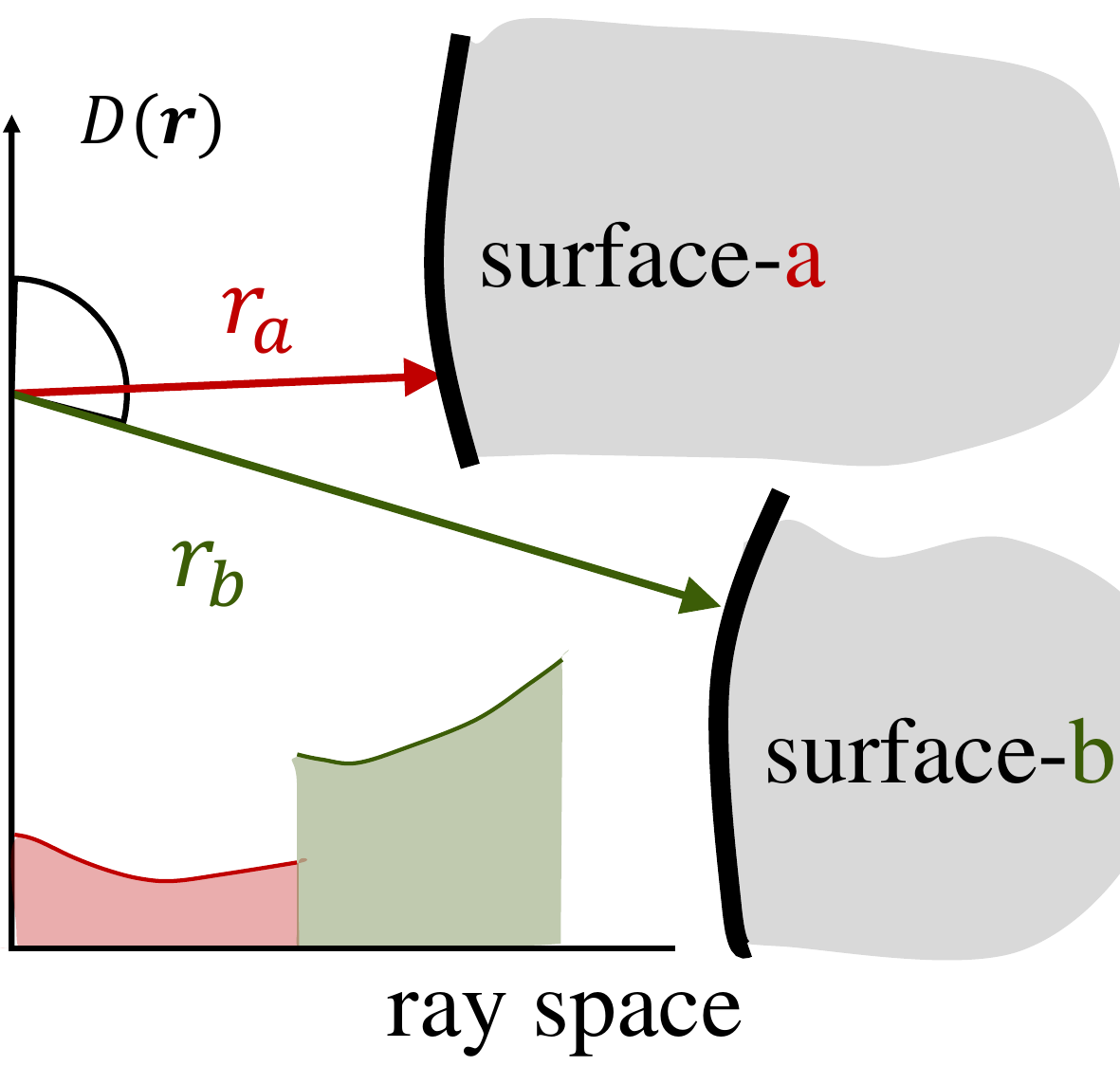}
  }
  \subfigure[]{
    \includegraphics[width=0.4\linewidth]{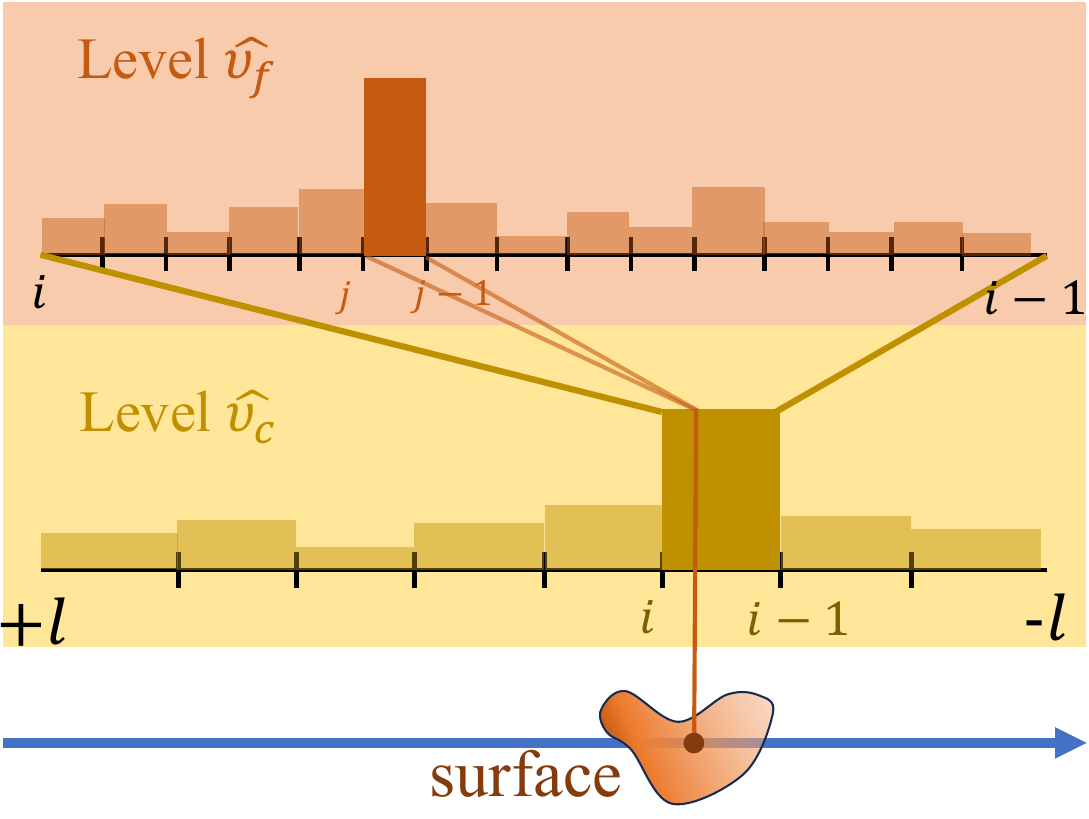}
  }
  \subfigure[]{
    \includegraphics[width=0.15\linewidth]{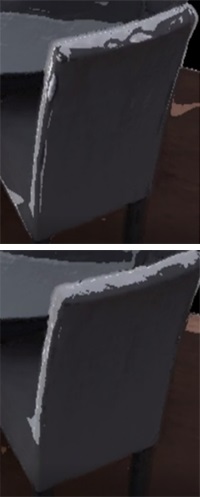}
  }
  \caption{Directly regressing $\mu$ may result in significant errors in the discontinuity area (a) during composition (see the top row of c). We address this problem by replacing the regression of $\mu$ with a multi-level classifier (b), which yields a better result (see the bottom row of c).}
  
  \label{fig:multi-classifier}
\end{figure}

Directly implementing the intersection network as a regression problem of the distance 
$\mu$ is not a good choice and tends to produce an inaccurate estimation because of the depth discontinuity issue \cite{neff2021donerf} during composition, as illustrated in Fig. \ref{fig:multi-classifier}(a).
This inaccurate estimation around edges can lead to floating artifacts during rendering as shown in Fig. \ref{fig:multi-classifier}(c). 

To address this problem, we instead build the intersection network as a multi-level classifier. As shown in Fig. \ref{fig:multi-classifier}(b), we first regularize $\mu$ in a limited range $[-l,l]$ with symmetric center as $\mathbf{p}_{\perp}$. We then divide the limited range into two levels using a segment operator $\mathcal{S}$, namely the coarse level with $N_c$ bins and the fine level with $N_f$ bins. The intersection network then is formulated as:
\begin{equation}
\mathcal{F}_{\Phi}:\mathbf{r} \mapsto (\hat{\upsilon}_c,\hat{\upsilon}_f,\hat{\alpha}),
\label{eq:NeDF_mulcla}
\end{equation}
where $\hat{\upsilon}_c,\hat{\upsilon}_f$ are 1D vectors representing the coarse- and fine-level target bins to which the surface belongs, respectively. The resolution of the two-level classifier is $N_c \times N_f$ ( where $N_c$ and $N_f$ are set to 64 and 128, respectively). Then, $\mu$ is calculated as follows:
\begin{equation}
    \mu = \mathcal{S}^{-1}\circ \mathcal{F}_{\Phi} =
    \lambda_1 p_c(\hat{\upsilon}_c)+ \lambda_2 p_f(\hat{\upsilon}_f) -l,
\end{equation}
where $\mathcal{S}^{-1}$ is the inverse operator of $\mathcal{S}$, $p_c(\hat{\upsilon}_c)=argmax(\hat{\upsilon}_c)/N_c$, $p_f(\hat{\upsilon}_f)=argmax(\hat{\upsilon}_f)/N_f$, $\lambda_1$ is $2l$ and $\lambda_2$ is $2l/N_c$. Because the target bin's indicator operation is a non-differentiable operation, we choose to use binary cross-entropy (BCE) loss (see Eq. \ref{eq:loss_nedf}) during training. We use the \textit{argmax} function to pick out the target bin during testing. 

In our implementation, the intersection network is an MLP. The MLP's head is a single linear layer; the MLP's body is made up of 16 residual blocks, each of which is made up of two fully connected layers followed by ReLU activation; the MLP's tail has two branches to produces $\hat{\alpha}$, $\hat{\upsilon}_c$ and $\hat{\upsilon}_f$ (see Appendix for more details). 
We supervise this intersection network with the pre-trained NeRF models. 
Specifically, the distance from the camera to the first ray-object intersection can be measured by:
\begin{equation}
    D(\mathbf{r})=\sum_{i=1}^N w_i \left(t_i |\mathbf{d}| \right), t_i \in [t_n,t_f],
    \label{eq:vol_depth}
\end{equation}
where $w_i$ is the weight of i-th sample along rays in NeRF's render equation, $\mathbf{d}$ is the direction of ray $\mathbf{r}$, $t_n$ and $t_f$ are the near and far clip ranges of the camera, respectively. This depth is then converted to $\mu$ using the inverse of Eq. \ref{eq:convert_depth} for supervision. 

The total training loss for the intersection network can be formulated as: 
\begin{equation}
\label{eq:loss_nedf}
    \mathcal{L}_{NeDF} = \text{BCE}(\hat{\upsilon}_c,\upsilon_c) + \text{BCE}(\hat{\upsilon}_f,\upsilon_f) + 0.1 \times \text{BCE}(\hat{\alpha},\alpha),
\end{equation}
where $\upsilon_c,\upsilon_f$ are one-hot vectors of the same dimension as $\hat{\upsilon}_c,\hat{\upsilon}_f$. They can be obtained by applying segment operator $\mathcal{S}$ on $\mu$ generated by NeRF. $\alpha$ is the ground truth mask generated by NeRF.

\subsection{Fast NeRF Composition with NeDF}
Built on NeDF, we are able to achieve a fast composition and rendering framework. We illustrate the structure and data processing flow of this framework on the figure in the right of Fig. \ref{fig:overview}.  It generally consists of three steps. First, the NeDF generation step takes all of the rays cast from the camera (eye position) through each pixel of the frame as input to output a depth, $D(\mathbf{r})$, and object ID buffers. It saves the corresponding depth and object identification information for each element of $D(\mathbf{r})$. 
Deferred Shading step, as the second step, takes the ID and depth buffers outputted by the previous NeDF Generation step and the rays $\mathbf{r}$ as an input. Then, it outputs a color buffer that holds the raw color (without shadow) information about the pixels. 
Lasty, the Shadow step uses the shadow rays, $D(\mathbf{r})$, and $\mathbf{r}$ to generate the shadow information in a shadow buffer output. The final rendering result is produced by multiplying the color buffer and shadow buffer. Next, we first describe how the method handles the affine transformation of each NeRF object, as well as the necessary details for rendering acceleration.  

\paragraph{Manipulating Objects.} Given a 3D vector $\mathbf{v}$ in world space and a corresponding 3D vector $\mathbf{q}$ in canonical space, we define the rigid transformation of a NeRF object as
 a mapping function $\mathcal{G}: \mathbf{q}(\mathbf{R}, \mathbf{T}, s) \mapsto \mathbf{v} $ where $\mathbf{R}, \mathbf{T}, s$ specify the rotation, translation, and scale, respectively.
Previous NeRF works focusing on deformation such as~\cite{Pumarola_2021_CVPR, park2021nerfies, park2021hypernerf, ost2021neural} transform the sampling points to canonical space to achieve the transformation. We instead directly transform the starting point and direction of rays to local canonical space and use this transformed rays to query the implicit field.
As the intersection network takes rays $\mathbf{r}$ as input, the depth $\hat{D}(\mathbf{r})$ in global space can be computed using the inverse of the affine transformation function $\mathcal{G}$:
\begin{equation}
    \hat{D}(\mathbf{r}) = |\mathbf{o}-\mathbf{p}_{\perp}|- s\left(\mathcal{S}^{-1} \circ \mathcal{F}_{\Phi}\right) \left(\mathcal{G}^{-1}(\mathbf{r})\right).
\end{equation}

\paragraph{Deferred Shading.} We term the step generating the raw color information as Deferred Shading step because
each pixel only requires one NeRF network $\mathcal{F}_{\Theta}$ to shade in our framework. As the intersection network can accurately estimate the depth information, we are able to replace the ray marching step in previous NeRF works with a degenerated ray casting form,
i.e., a pixel color can be acquired by a single evaluation of $\mathcal{F}_{\Theta}$. To further shade all objects in a lighting-aware manner, it is necessary to handle different lighting conditions in the original space from which the objects are collected, which is a non-trivial task. We simplify this processing by only shading with the primary color. Specifically, we obtain each ray's color by:
\begin{equation}
    \mathbf{c},\sigma = \mathcal{F}_{\Theta}\left(\mathcal{G}^{-1}(\mathbf{o}+\hat{D}(\mathbf{r})\mathbf{d}),
    \mathcal{G}^{-1}(\mathbf{d})\right),
    \label{eq:deferred_color}
\end{equation}
where color $c$ at the ray-object intersection point is directly used to approximate the ray's color $\hat{C}(\mathbf{r})$. Density $\sigma$ is employed to check whether the estimated $\hat{D}$ is convincible. An extremely small $\sigma$ indicates that $\hat{D}$ does not reach the object surface. In this situation, we set a threshold that optionally assists in resampling these outliers using primary volume render equation in NeRF for better quality but a slight speed penalty (see our experimental results in Table \ref{tab:bxobj}).
Although replacing the raymarching step with a single sample limits its usefulness, it is adequate for previewing.
 Users may prefer interactive-level speed over perfect quality when manipulating each object. In addition, once the adjustment is confirmed, the pipeline can re-render the scene and generate a set of high-quality results using only the NeRF models.

\begin{figure*}[htbp]
  \centering
  \subfigure[NeDF+Mip-NeRF]{
    \includegraphics[width=0.32\linewidth]{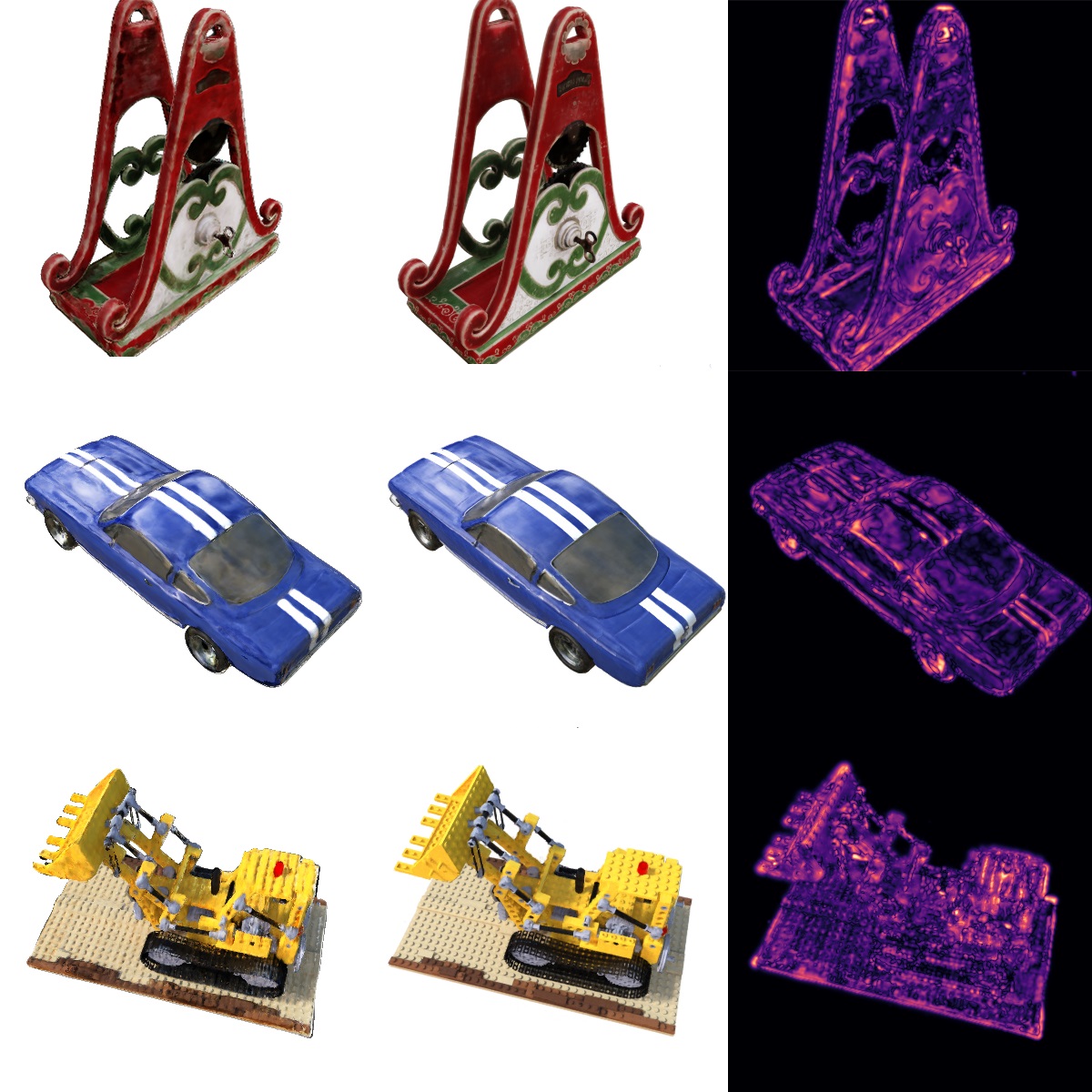}  
  }
  \subfigure[NeDF+SNeRF]{
      \includegraphics[width=0.32\linewidth]{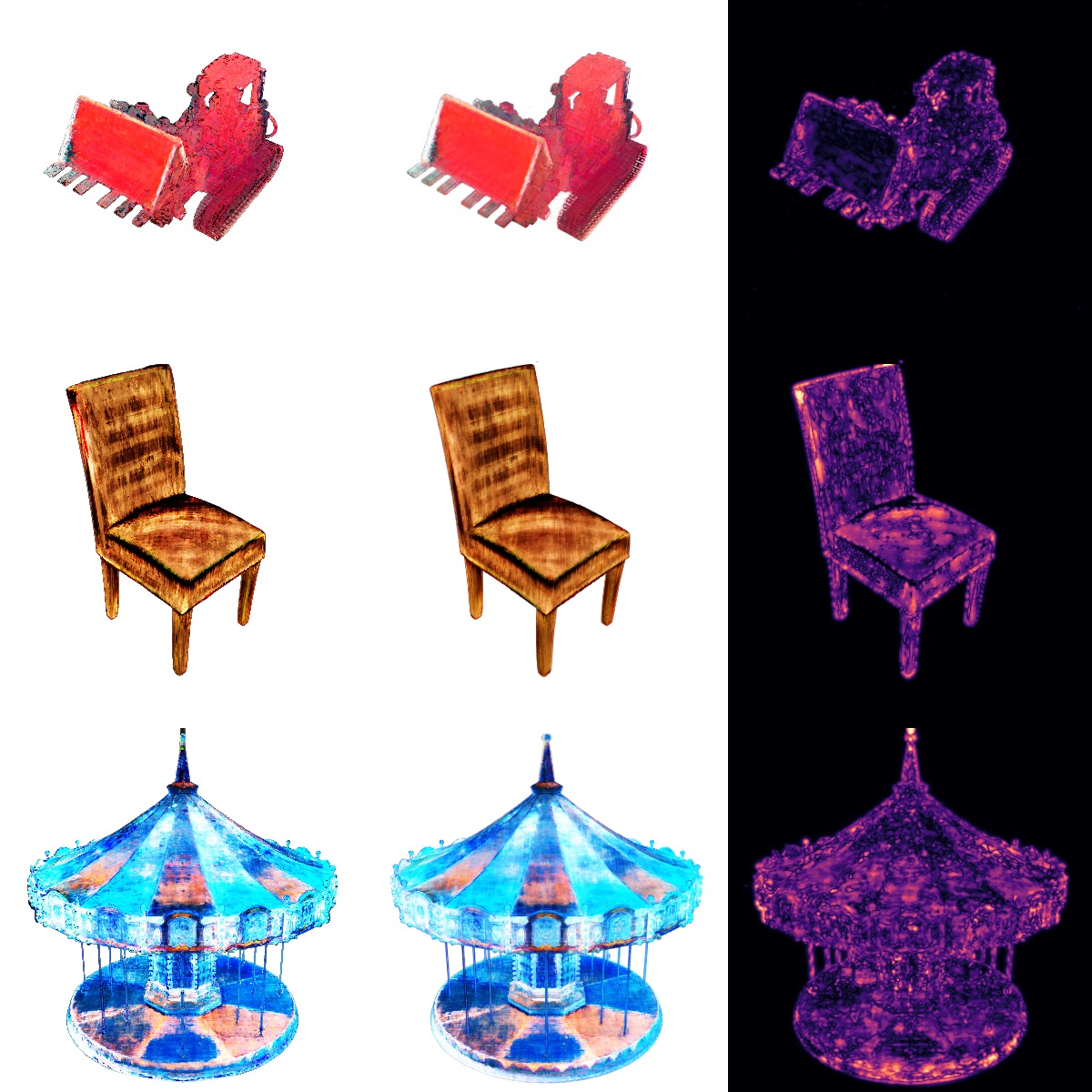}
  }
  \subfigure[NeDF+Neus]{
      \includegraphics[width=0.32\linewidth]{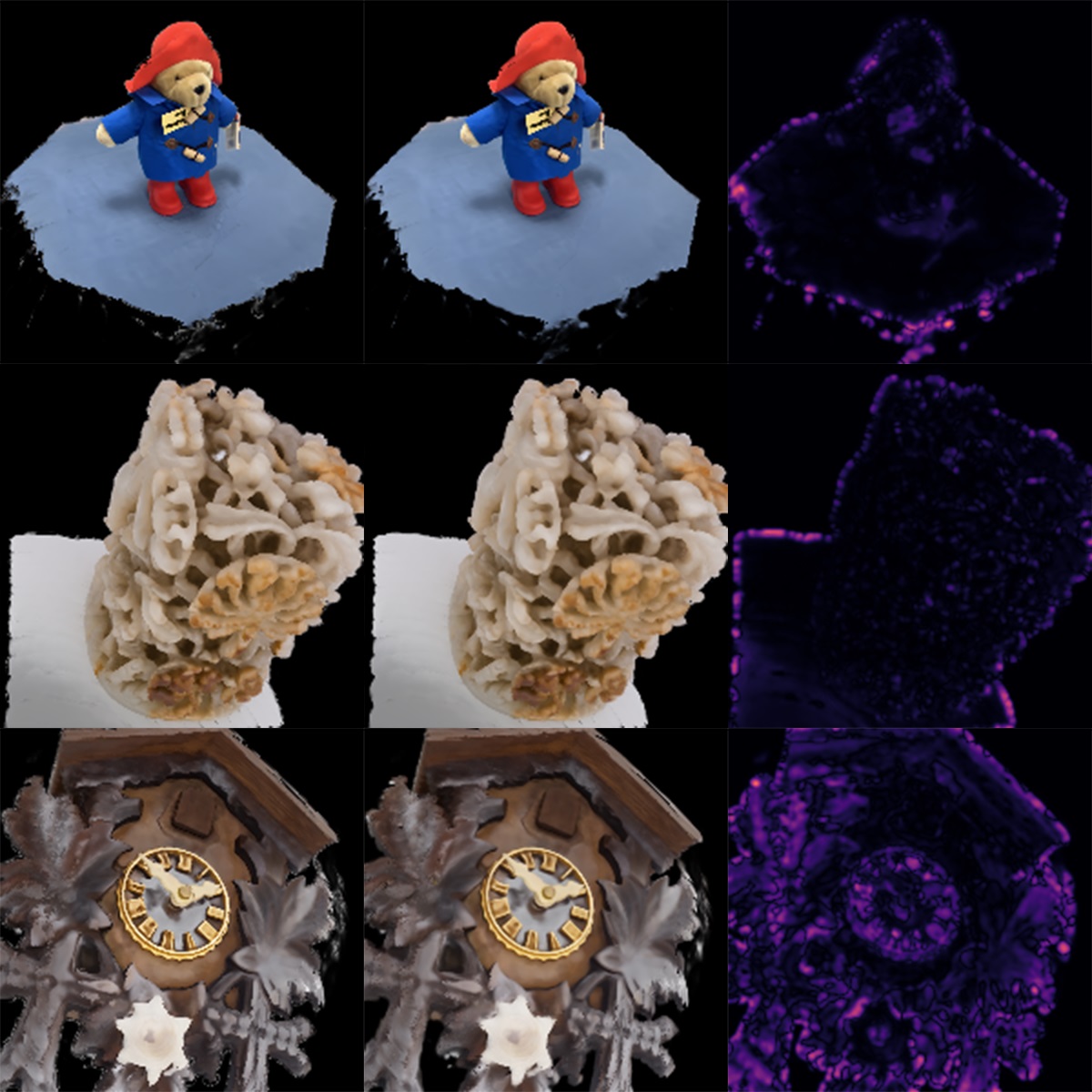}
  }
  \caption{Visual quality illustration of the general plugin. For each method, the left column shows the previewing result, the middle column shows the naive result, and the right column visualizes the flip errors \cite{Andersson2020}.}
  \label{fig:3based}
\end{figure*}

\begin{table*}[htbp]
\begin{center}
\begin{tabular}{c|cc|cc|cc}
\hline
 & \multicolumn{2}{c|}{NeDF+Mip-NeRF vs. Mip-NeRF } & \multicolumn{2}{c|}{NeDF+SNeRF vs. SNeRF} & \multicolumn{2}{c}{NeDF+Neus vs. Neus} \\ \cline{2-7} 
 & time (s)   & s-time(\%)$\uparrow$   & time (s)   & s-time(\%)    &  time (s)  & s-time(\%)   \\ \hline
Outlier Processing (rs)    &0.417/15.677  & 97.34 & 0.519/15.528 & 96.66 & 2.11/21.72  & 90.29 \\
w/o Outlier Processing (wo rs) &0.234/15.677  & 98.51 & 0.225/15.528  & 98.55 & 0.202/21.72  & 99.07 \\ \hline
\end{tabular}
\caption{Our proposed framework serves as a plugin for three representative NeRF models: Mip-NeRF, SNeRF, and Neus. Term "s-time" refers to the percentage of time that our method saves. The optional outlier processing in Eq. \ref{eq:deferred_color} provides better quality but a slight speed penalty. The resolution is set to $800\times800$.} 
\label{tab:3based}
\end{center}
\end{table*}


\subsection{Fast Dynamic shadows Casting}
Shadows for NeRF models can be generated using the traditional volume rendering technique developed by \cite{max1995optical}. It, however, requires an additional volumetric evaluation for each sample point along a ray to determine whether the point is in shadows, which significantly increases the time consumption. 
Given that target objects have opaque in our cases, we generate the dynamic shadows based on the fundamental concept of shadow mapping \cite{williams1978casting} with NeDF for quick preview. We detail this dynamic shadow casting as an additional step (Shadow Step, also named step 3) in appendix.

\section{Experiments}
In this section, we first demonstrate that our pipeline indeed takes effect for various NeRF works. Following that, we show how our framework can assemble a large number of neural objects to form a virtual 3D world. Furthermore, we show our approach can be combined with traditional rendering techniques to create a mixed rendering pipeline. We implement our framework in two versions: Pytorch version for a convenient reproduction and CUDA version for testing performance limitation. We conduct experiments on our new $N$-object dataset which has been described in section '$N$ -object dataset testing'.
\paragraph{Training details.}A single Nvidia A100 GPU is used to train NeDF for each scene from our $N$-object dataset. We employ Adam as the optimizer and set the learning rate to 5e-4. We use the pre-trained NeRF models to generate 500 random views (more random views, less artifacts) for supervision for each NeDF model of a single object, and train the intersection network over 60W iterations until convergence. The batchsize of training rays is set to 4,096 during each iteration.

\subsection{General Acceleration Plugin}
To evaluate the efficacy of the proposed framework acting as a general pipepline, we conduct experiments on the pre-trained models of certain classic implicit NeRFs. Three representative
implicit neural representation based NeRF methods, i.e., Mip-NeRF \cite{barron2021mip},
SNeRF \cite{snerf2022}, and Neus \cite{wang2021neus}, are employed for this set of experiments. 
Mip-NeRF represents the kind of NeRF methods that focus on novel view synthesis and rendering. SNeRF is a typical NeRF method that supports appearance style transfer, while Neus is a representative for NeRF methods designed for task of geometry reconstruction.
We integrate these three representative NeRF techniques on our dataset and show their corresponding quantitative performance results in Table \ref{tab:3based} and visual results in Fig. \ref{fig:3based}. Evidently, we can see that these representative NeRF methods, when used in conjunction with our proposed framework, significantly reduce rendering time while maintaining an acceptable level of quality degradation.

\subsection{Storage and Speed}
As shown in SNeRF (the same architecture with vanilla NeRF) column in Table \ref{tab:3based}, our framework can speed up rendering more than $30 \sim 40$ times over the vanilla NeRF.
\paragraph{Comparision with other solutions.} Although there's no general pipeline for NeRF composition, we force a comparsion with NSVF~\cite{liu2020neural}, the closest competitor for
NeRF fast composition, and some other works focus on storage in compositing. In terms of speed, it has demonstrated that our framework can save more than twice as much time as NSVF ($\sim 20 \times$, see Figure 2 in the supplementary appendix). In terms of storage, our
models (NeDF+NeRF) requires 13MB (3MB for NeRF), NSVF (grid+NeRF) requires 3.2$\sim$16MB, DONeRF requires 3.6MB, and CC-NeRF requires 1.5MB. It should be noted that CC-NeRF is not intended for fast composition, and DONeRF is incapable of
handling arbitrary composition with affine transformation.
By comparing our method to these, we demonstrate that our method does not require excessive storage and is sufficiently
fast. What's more, our framework has several significant advantages that the other approaches (NSVF and CC-NeRF) lack, including:
1) unlike to our framework, neither NSVF nor CC-NeRF can be used as a general acceleration plugin
and 2) NSVF and CC-NeRF cannot achieve interactive-level manipulation speeds or generate shadows like our method and 3) both of CC-NeRF and NSVF incur additional memory cost when rendering repeated objects while our framework avoids this issue.

\subsection{Capbility evalution}
\paragraph{$N$-object dataset testing.}
\begin{figure}[tbp]
    \centering
    \includegraphics[width=\linewidth]{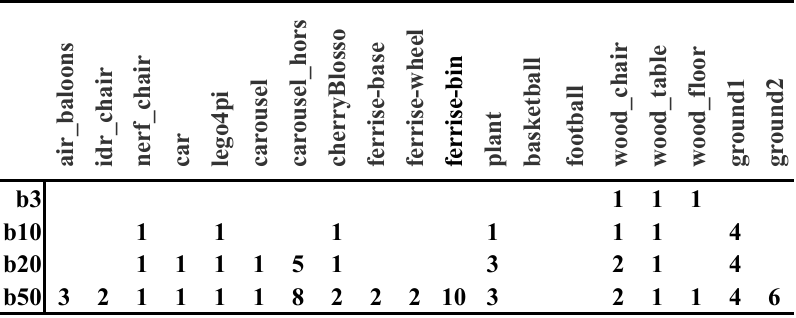}
    \caption{Number of objects contained in \textit{b3},\textit{b10},\textit{b20},\textit{b50} scene. The number in two ball cases is empty, which is because we use them for the capability pressure test and repeated basketballs and footballs (sharing fifty-fifty) in the 100-ball and 1k-ball scenes.}
    \label{fig:bx_table}
\end{figure}

\begin{figure}[tbp]
  \centering
  \begin{minipage}{\linewidth}
      \includegraphics[width=\linewidth]{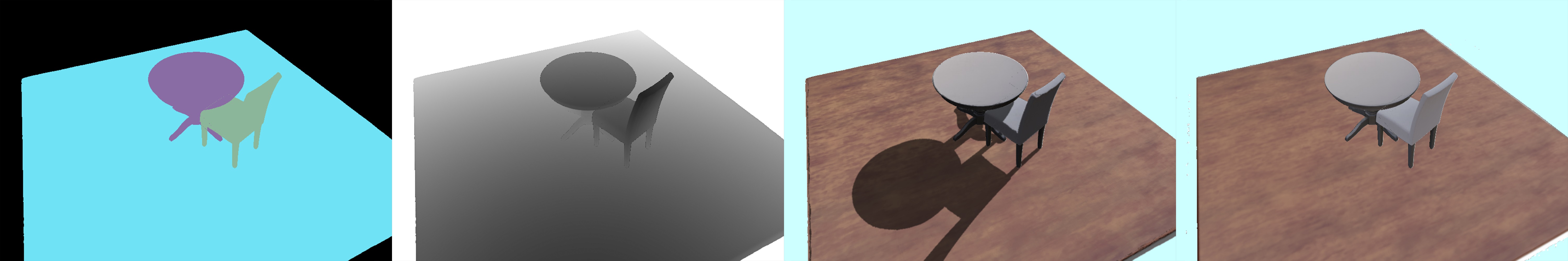} 
      \includegraphics[width=\linewidth]{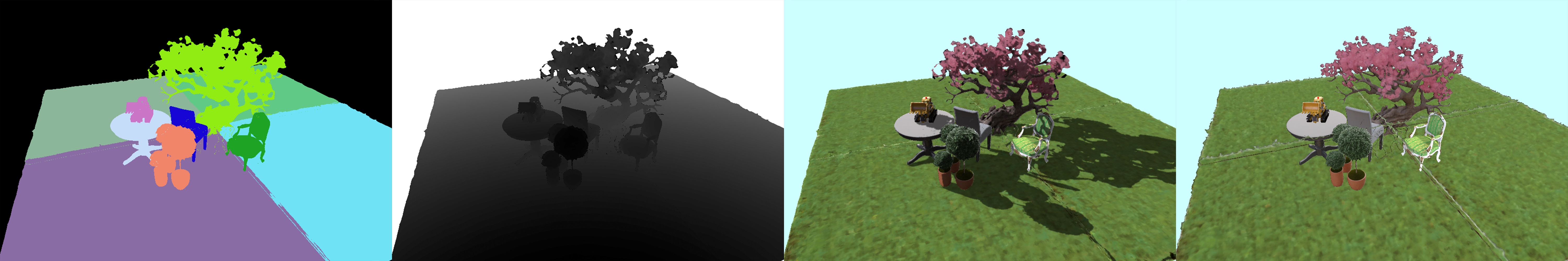} 
      \includegraphics[width=\linewidth]{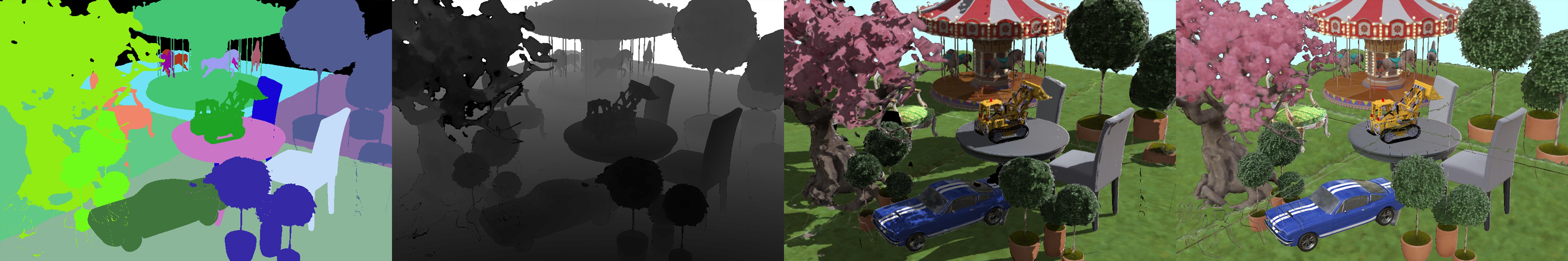} 
      \includegraphics[width=\linewidth]{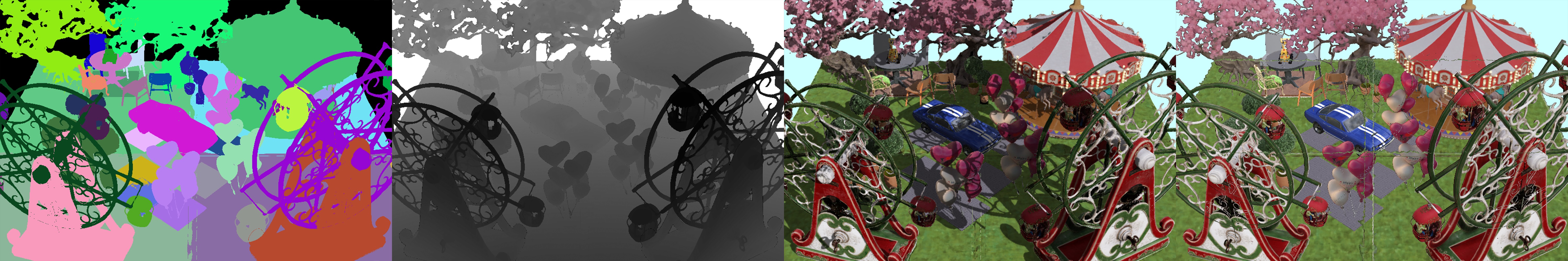} 
      \includegraphics[width=\linewidth]{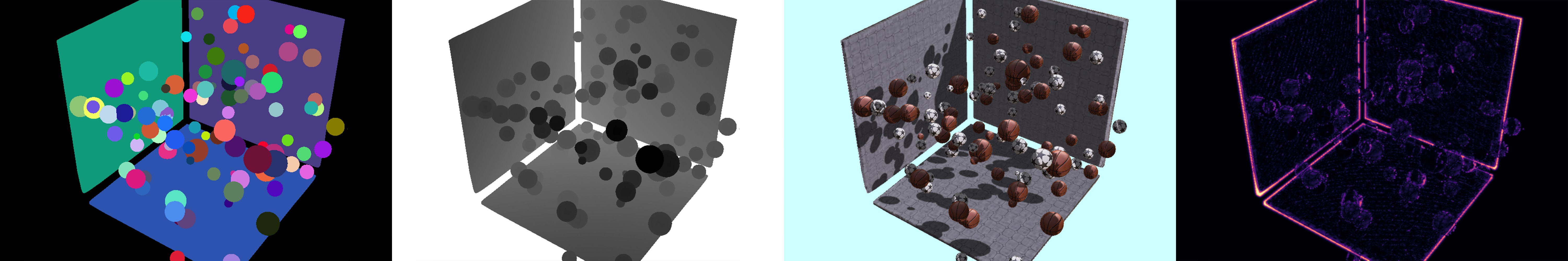}
      \includegraphics[width=\linewidth]{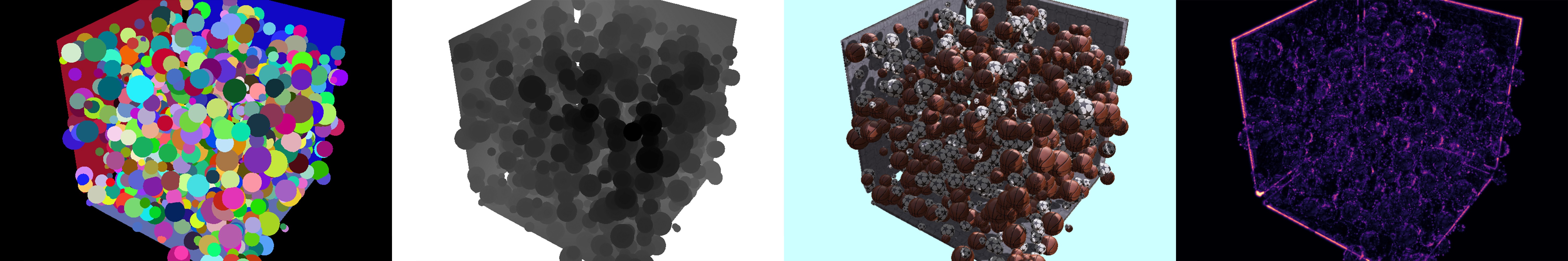}
  \end{minipage}
  \caption{Results on our $N$-object dataset. Each 4-image pair shows: ID map, depth map, results with shadows using our previewing framework, and results with only naive NeRFs (shadowless). The Rightside of last two rows is the flip error \cite{Andersson2020} between previewing and NeRF's naive composition results). 
  From top to bottom: \textit{b3 and b10, b20 and b50, 100-ball and 1k-ball} virtual scenes. Scenes \textit{b20} and \textit{b50} are both dynamic, with carousels and ferris wheels moving through user control settings; scenes \textit{100-ball} and \textit{1k-ball} have 100 and 1,000 balls (half basketballs and half footballs), respectively.
  }
  \label{fig:bx-obj}
\end{figure}

First, we collected an $N$-object dataset consisting of 22 distinct NeRF objects: 7 objects were chosen from previous work (3 from Neus' dataset of real-world objects, 4 from synthesis dataset \cite{mildenhall2020nerf,zhang2022modeling,verbin2022ref}), and the remaining 15 objects were newly created by Blender 3D. To evaluate the framework's ability to composite complex virtual scenes (see Fig. \ref{fig:bx-obj}), we
conduct a set of virtual scene compositions with four-level increasing complexity: \textit{b3, b10, b20, b50}.
The \textit{b3} and \textit{b10} are two static scenes with lower complexity, and the \textit{b20} and \textit{b50} are dynamic scenes where some objects move according to user-defined transformation settings. We allow to insert the same NeRF object instances repeatedly into the scene to increase the complexity of the virtual scene.
For each scene, we set different camera views (such as the views in Fig. \ref{fig:bx-obj}) and render 120 frames to evaluate the average performance. Table \ref{tab:bxobj} shows the consumed time and quality of each composition.
Figure \ref{fig:step_ratio} depicts the time consumption ratio of each rendering step (see appendix) in rendering a single frame. As complexity increases the resampling ratio increases, which results in significant differences in computational time across different scenes in the \textit{Deferred Shading step}.
To test the capability upper bound of our framework, we also conduct a pressure test on virtual scenes with 100 and 1k balls to see how stable the proposed approach is for scene composition and shadow casting (see Fig. \ref{fig:bx-obj}). Despite taking a longer time (see Tab. \ref{tab:bxobj}), our pipeline delivers stable performance and does not incur additional memory cost for those repeated balls.

\begin{table*}[tbp]
\begin{tabular}{ll|ll|ll|ll|ll|ll|ll}
\hline
 &
 &
  \multicolumn{2}{c|}{b3-scene} &
  \multicolumn{2}{c|}{b10-scene} &
  \multicolumn{2}{c|}{b20-scene} &
  \multicolumn{2}{c|}{b50-scene} &
  \multicolumn{2}{c|}{100-ball} &
  \multicolumn{2}{c}{1k-ball} \\ \hline
                             &        & time(s) & PSNR & time & PSNR & time & PSNR & time & PSNR & time & PSNR & time & PSNR \\ \hline
\multirow{2}{*}{wo s}        & rs & 0.240    & 25.78       & 0.658    & 23.32       & 1.678    & 20.02      & 1.881    & 18.67       & 1.887    & 26.54   & 16.63    & 25.26       \\
                             & wo rs   & 0.209    & 23.97    & 0.443    & 19.92   & 0.873    & 15.26   & 1.209    &  15.43  & 1.741    & 24.19   & 16.34    & 22.58       \\
\multirow{2}{*}{shadow} & rs & 0.398    & \multirow{2}{*}{\textbackslash}       & 1.040    & \multirow{2}{*}{\textbackslash}       & 2.441    & \multirow{2}{*}{\textbackslash}       & 2.943    & \multirow{2}{*}{\textbackslash}       & 3.073    & \multirow{2}{*}{\textbackslash}       & 28.97    & \multirow{2}{*}{\textbackslash}       \\
                             & wo rs   & 0.367    &        & 0.813    &        & 1.628    &        & 2.286    &        & 3.002    &       & 27.66  &        \\
\hline
\end{tabular}
\caption{Performance of the implemented Pytorch framework on an $N$-object dataset ('wo s' means no shadow is calculated; 'rs' stands for resampling, i.e., outlier processing in Table~1). The resolution is set to $900\times600$. We use the naive composition (right side in Fig. \ref{fig:bx-obj}) by NeRF as ground truth for composition without shadows. We only test the average time (in seconds) in each frame for shadow composition. It should be noted that the quality of the 1K-ball pressure testing was superior to the former due to the simpler geometry of the objects. Nonetheless, due to excessive overlapping between balls, the time increases rapidly.}
\label{tab:bxobj}
\end{table*}

\begin{figure}[htbp]
    \centering
    \includegraphics[width=\linewidth]{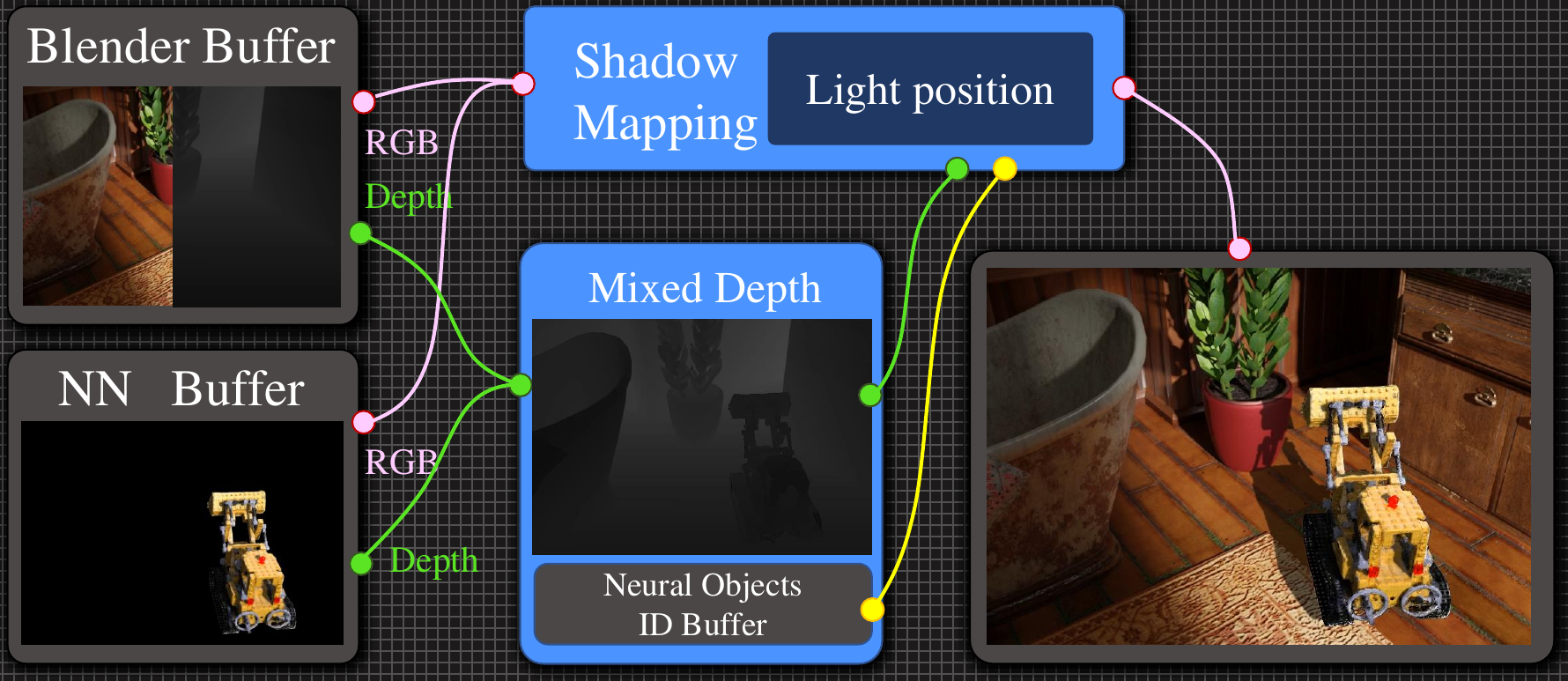}
    \caption{Given the information of camera view and light position, the buffer produced by the proposed framework (NN Buffer) and the \textit{Cycles} engine of Blender 3D can be combined to form a mixed render result (see our video for a dynamic result). The \textit{Bathroom} scene is from \cite{huang2022hdr}.}
    \label{fig:mixed_pipeline}
\end{figure}

\begin{figure}[htbp]
    \centering
    \includegraphics[width=\linewidth]{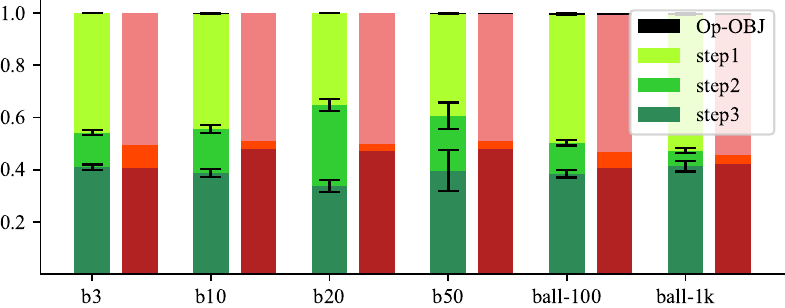}
    \caption{The time consumption ratio of our framework for the main steps. 
    \textit{Green} and \textit{Red} bars represent the results with and without outliers resampling, respectively. Regarding the green bar, refer to this figure in conjunction with Fig. \ref{fig:bx-obj}, when the camera view varies, certain objects which occupy a larger portion of the frame, e.g., the trees, have a higher resampling ratio and thus increase the processing time. Focusing attention on the red bar, the Step 2 (deferred shading step) consistently assumes a relatively minor temporal allocation across all scenarios. This is because, regardless of the number of objects in the scene, a pixel will be shaded only once with a single NeRF object.
    }
    \label{fig:step_ratio}
\end{figure}

\paragraph{CUDA implementation.} 
CUDA version of the proposed framework is built with Vulkan, TensorRT, and a customized CUDA kernel to enable real-time manipulation on Nvidia RTX 4090 GPU. With this, we enable the user to construct a virtual scene in a progressive and interactive manner.
The real-time manipulation benefits from a buffer reuse technique to accelerate the performance during manipulation. Specifically, when the user manipulates an object, the other objects' rendering processes are frozen, and their buffers are reused to decrease the computational load.

\subsection{Compatibility with traditional graphics pipelines} 
The proposed framework is also suitable for integration with traditional rendering pipelines to build a mixed pipeline that supports dynamic shadow casting across meshes and neural objects in a single scene (see Fig. \ref{fig:mixed_pipeline}). Specifically, we first use a shared camera \& light settings across mesh and neural scene to generate rendering buffers. Subsequently, these two scenes are seamlessly integrated using depth-based composition. Finally, shadow mapping is applied to produced a mixed result.

\section{Conclusions and Limitations}
We present a general framework for a fast composition of NeRF objects with dynamic shadows. The core technical contribution is the use of Neural Depth Fields to allow for the direct intersection of rays and implicit surfaces to quickly determine the spatial relationship between objects as well as surface color computation of NeRF objects. To the best of our knowledge, this is the first framework that allows for the rapid creation of a virtual scene composed of a large number of NeRF objects with dynamic shadows. Our method can also be combined with traditional renderers to form a mixed pipeline, increasing its versatility and applicability.

The proposed framework is currently limited to solid surface objects, and predicting weight distribution in local space rather than a single depth may be a better option. 
In the future, this compositing framework can be further enhanced with more powerful editable NeRF works \cite{martinbrualla2020nerfw, tancik2022block,snerf2022, tojo2022posternerf,zhang2021nerfactor, zhang2022modeling, verbin2022ref} to achieve global illumination and improved visual quality.

\section{Acknowledgments}
This work was supported by Key R\&D Program of Zhejiang (No. 2023C01047) and the National Natural Science Foundation of China (Grant No. 62036010). The authors acknowledge the support of Bytedance's MM Lab's GPU cluster, as well as Yanzhen Chen's early assistance in searching for various works. 

\bibliography{aaai24}

\begin{thebibliography}{50}
\providecommand{\natexlab}[1]{#1}

\bibitem[{Andersson et~al.(2020)Andersson, Nilsson, Akenine{-}M{\"{o}}ller, Oskarsson, {\AA}str{\"{o}}m, and Fairchild}]{Andersson2020}
Andersson, P.; Nilsson, J.; Akenine{-}M{\"{o}}ller, T.; Oskarsson, M.; {\AA}str{\"{o}}m, K.; and Fairchild, M.~D. 2020.
\newblock {{\FLIP:} {A} difference evaluator for alternating images}.
\newblock \emph{Proceedings of the ACM on Computer Graphics and Interactive Techniques}, 3(2): 15:1--15:23.

\bibitem[{Attal et~al.(2022)Attal, Huang, Zollh{\"o}fer, Kopf, and Kim}]{attal2022learning}
Attal, B.; Huang, J.-B.; Zollh{\"o}fer, M.; Kopf, J.; and Kim, C. 2022.
\newblock Learning neural light fields with ray-space embedding networks.
\newblock In \emph{Proceedings of the IEEE/CVF Conference on Computer Vision and Pattern Recognition (CVPR)}, 19819--19829.

\bibitem[{Barron et~al.(2021)Barron, Mildenhall, Tancik, Hedman, Martin-Brualla, and Srinivasan}]{barron2021mip}
Barron, J.~T.; Mildenhall, B.; Tancik, M.; Hedman, P.; Martin-Brualla, R.; and Srinivasan, P.~P. 2021.
\newblock Mip-nerf: A multiscale representation for anti-aliasing neural radiance fields.
\newblock In \emph{Proceedings of the IEEE/CVF International Conference on Computer Vision}, 5855--5864.

\bibitem[{Boss et~al.(2021{\natexlab{a}})Boss, Braun, Jampani, Barron, Liu, and Lensch}]{boss2021nerd}
Boss, M.; Braun, R.; Jampani, V.; Barron, J.~T.; Liu, C.; and Lensch, H. 2021{\natexlab{a}}.
\newblock Nerd: Neural reflectance decomposition from image collections.
\newblock In \emph{Proceedings of the IEEE/CVF International Conference on Computer Vision}, 12684--12694.

\bibitem[{Boss et~al.(2021{\natexlab{b}})Boss, Jampani, Braun, Liu, Barron, and Lensch}]{boss2021neural}
Boss, M.; Jampani, V.; Braun, R.; Liu, C.; Barron, J.; and Lensch, H. 2021{\natexlab{b}}.
\newblock Neural-pil: Neural pre-integrated lighting for reflectance decomposition.
\newblock \emph{Advances in Neural Information Processing Systems}, 34: 10691--10704.

\bibitem[{Cao et~al.(2022)Cao, Wang, Chemerys, Shakhrai, Hu, Fu, Makoviichuk, Tulyakov, and Ren}]{cao2022real}
Cao, J.; Wang, H.; Chemerys, P.; Shakhrai, V.; Hu, J.; Fu, Y.; Makoviichuk, D.; Tulyakov, S.; and Ren, J. 2022.
\newblock Real-time neural light field on mobile devices.
\newblock \emph{arXiv preprint arXiv:2212.08057}.

\bibitem[{Chen et~al.(2023)Chen, Funkhouser, Hedman, and Tagliasacchi}]{chen2022mobilenerf}
Chen, Z.; Funkhouser, T.; Hedman, P.; and Tagliasacchi, A. 2023.
\newblock MobileNeRF: Exploiting the Polygon Rasterization Pipeline for Efficient Neural Field Rendering on Mobile Architectures.
\newblock In \emph{The Conference on Computer Vision and Pattern Recognition (CVPR)}.

\bibitem[{Dansereau and Bruton(2004)}]{1328805}
Dansereau, D.; and Bruton, L. 2004.
\newblock Gradient-based depth estimation from 4D light fields.
\newblock In \emph{2004 IEEE International Symposium on Circuits and Systems (IEEE Cat. No.04CH37512)}, volume~3, III--549.

\bibitem[{Gu et~al.(2022)Gu, Liu, Wang, and Theobalt}]{gu2021stylenerf}
Gu, J.; Liu, L.; Wang, P.; and Theobalt, C. 2022.
\newblock StyleNeRF: A style-based 3D aware generator for high-resolution image synthesis.
\newblock In \emph{International Conference on Learning Representations}.

\bibitem[{Hedman et~al.(2021)Hedman, Srinivasan, Mildenhall, Barron, and Debevec}]{hedman2021baking}
Hedman, P.; Srinivasan, P.~P.; Mildenhall, B.; Barron, J.~T.; and Debevec, P. 2021.
\newblock Baking neural radiance fields for real-time view synthesis.
\newblock In \emph{Proceedings of the IEEE/CVF International Conference on Computer Vision}, 5875--5884.

\bibitem[{Huang et~al.(2022)Huang, Zhang, Feng, Li, Wang, and Wang}]{huang2022hdr}
Huang, X.; Zhang, Q.; Feng, Y.; Li, H.; Wang, X.; and Wang, Q. 2022.
\newblock Hdr-nerf: High dynamic range neural radiance fields.
\newblock In \emph{Proceedings of the IEEE/CVF Conference on Computer Vision and Pattern Recognition}, 18398--18408.

\bibitem[{Jiang et~al.(2022)Jiang, Chen, Liu, Fu, and Gao}]{jiang2022nerffaceediting}
Jiang, K.; Chen, S.-Y.; Liu, F.-L.; Fu, H.; and Gao, L. 2022.
\newblock NeRFFaceEditing: Disentangled face editing in neural radiance fields.
\newblock In \emph{SIGGRAPH Asia 2022 Conference Papers}, 1--9.

\bibitem[{Kim et~al.(2013)Kim, Zimmer, Pritch, Sorkine-Hornung, and Gross}]{Kim2013scene}
Kim, C.; Zimmer, H.; Pritch, Y.; Sorkine-Hornung, A.; and Gross, M. 2013.
\newblock Scene reconstruction from high spatio-angular resolution light fields.
\newblock \emph{ACM Transactions on Graphics (Proceedings of ACM SIGGRAPH)}, 32(4): 73:1--73:12.

\bibitem[{Krishnan et~al.(2023)Krishnan, Raj, Zhang, Carlson, Tseng, Sridhar, Jaipuria, and Hays}]{krishnan2023lane}
Krishnan, A.; Raj, A.; Zhang, X.; Carlson, A.; Tseng, N.; Sridhar, S.; Jaipuria, N.; and Hays, J. 2023.
\newblock LANe: Lighting-aware neural fields for compositional scene synthesis.
\newblock arXiv:2304.03280.

\bibitem[{Kundu et~al.(2022)Kundu, Genova, Yin, Fathi, Pantofaru, Guibas, Tagliasacchi, Dellaert, and Funkhouser}]{kundu2022panoptic}
Kundu, A.; Genova, K.; Yin, X.; Fathi, A.; Pantofaru, C.; Guibas, L.~J.; Tagliasacchi, A.; Dellaert, F.; and Funkhouser, T. 2022.
\newblock Panoptic neural fields: A semantic object-aware neural scene representation.
\newblock In \emph{Proceedings of the IEEE/CVF Conference on Computer Vision and Pattern Recognition}, 12871--12881.

\bibitem[{Kurz et~al.(2022)Kurz, Neff, Lv, Zollh{\"o}fer, and Steinberger}]{kurz2022adanerf}
Kurz, A.; Neff, T.; Lv, Z.; Zollh{\"o}fer, M.; and Steinberger, M. 2022.
\newblock AdaNeRF: Adaptive sampling for real-Time rendering of neural radiance fields.
\newblock In \emph{European Conference on Computer Vision}, 254--270. Springer.

\bibitem[{Liu et~al.(2020)Liu, Gu, Zaw~Lin, Chua, and Theobalt}]{liu2020neural}
Liu, L.; Gu, J.; Zaw~Lin, K.; Chua, T.-S.; and Theobalt, C. 2020.
\newblock Neural sparse voxel fields.
\newblock \emph{Advances in Neural Information Processing Systems}, 33: 15651--15663.

\bibitem[{Ma et~al.(2023)Ma, Li, He, Dong, and Tan}]{Ma_Li_He_Dong_Tan_2023}
Ma, T.; Li, B.; He, Q.; Dong, J.; and Tan, T. 2023.
\newblock Semantic 3D-Aware Portrait Synthesis and Manipulation Based on Compositional Neural Radiance Field.
\newblock \emph{Proceedings of the AAAI Conference on Artificial Intelligence}, 37(2): 1878--1886.

\bibitem[{Martin-Brualla et~al.(2021)Martin-Brualla, Radwan, Sajjadi, Barron, Dosovitskiy, and Duckworth}]{martinbrualla2020nerfw}
Martin-Brualla, R.; Radwan, N.; Sajjadi, M.~S.; Barron, J.~T.; Dosovitskiy, A.; and Duckworth, D. 2021.
\newblock Nerf in the wild: Neural radiance fields for unconstrained photo collections.
\newblock In \emph{Proceedings of the IEEE/CVF Conference on Computer Vision and Pattern Recognition}, 7210--7219.

\bibitem[{Max(1995)}]{max1995optical}
Max, N. 1995.
\newblock Optical models for direct volume rendering.
\newblock \emph{IEEE Transactions on Visualization and Computer Graphics}, 1(2): 99--108.

\bibitem[{Meng et~al.(2019)Meng, So, Sun, and Lam}]{meng2019high}
Meng, N.; So, H. K.-H.; Sun, X.; and Lam, E.~Y. 2019.
\newblock High-dimensional dense residual convolutional neural network for light field reconstruction.
\newblock \emph{IEEE Transactions on Pattern Analysis and Machine Intelligence}, 43(3): 873--886.

\bibitem[{Mildenhall et~al.(2020)Mildenhall, Srinivasan, Tancik, Barron, Ramamoorthi, and Ng}]{mildenhall2020nerf}
Mildenhall, B.; Srinivasan, P.~P.; Tancik, M.; Barron, J.~T.; Ramamoorthi, R.; and Ng, R. 2020.
\newblock NeRF: Representing scenes as neural radiance fields for view synthesis.
\newblock In \emph{ECCV}, 405--421.

\bibitem[{M\"uller et~al.(2022)M\"uller, Evans, Schied, and Keller}]{mueller2022instant}
M\"uller, T.; Evans, A.; Schied, C.; and Keller, A. 2022.
\newblock Instant neural graphics primitives with a multiresolution hash encoding.
\newblock \emph{ACM Transactions on Graphics}, 41(4): 102:1--102:15.

\bibitem[{Neff et~al.(2021)Neff, Stadlbauer, Parger, Kurz, Mueller, Chaitanya, Kaplanyan, and Steinberger}]{neff2021donerf}
Neff, T.; Stadlbauer, P.; Parger, M.; Kurz, A.; Mueller, J.~H.; Chaitanya, C. R.~A.; Kaplanyan, A.; and Steinberger, M. 2021.
\newblock DONeRF: Towards real-time rendering of compact neural radiance fields using depth oracle networks.
\newblock \emph{Computer Graphics Forum}, 40(4): 45--59.

\bibitem[{Nguyen-Phuoc, Liu, and Xiao(2022)}]{snerf2022}
Nguyen-Phuoc, T.; Liu, F.; and Xiao, L. 2022.
\newblock SNeRF: Stylized Neural Implicit Representations for 3D Scenes.
\newblock \emph{ACM Transactions on Graphics}, 41(4).

\bibitem[{Niemeyer et~al.(2022)Niemeyer, Barron, Mildenhall, Sajjadi, Geiger, and Radwan}]{Niemeyer2021Regnerf}
Niemeyer, M.; Barron, J.~T.; Mildenhall, B.; Sajjadi, M. S.~M.; Geiger, A.; and Radwan, N. 2022.
\newblock RegNeRF: Regularizing Neural Radiance Fields for View Synthesis from Sparse Inputs.
\newblock In \emph{Proc. IEEE Conf. on Computer Vision and Pattern Recognition (CVPR)}, 5480--5490.

\bibitem[{Ost et~al.(2021)Ost, Mannan, Thuerey, Knodt, and Heide}]{ost2021neural}
Ost, J.; Mannan, F.; Thuerey, N.; Knodt, J.; and Heide, F. 2021.
\newblock Neural scene graphs for dynamic scenes.
\newblock In \emph{Proceedings of the IEEE/CVF Conference on Computer Vision and Pattern Recognition}, 2856--2865.

\bibitem[{Park et~al.(2021{\natexlab{a}})Park, Sinha, Barron, Bouaziz, Goldman, Seitz, and Martin-Brualla}]{park2021nerfies}
Park, K.; Sinha, U.; Barron, J.~T.; Bouaziz, S.; Goldman, D.~B.; Seitz, S.~M.; and Martin-Brualla, R. 2021{\natexlab{a}}.
\newblock Nerfies: Deformable neural radiance fields.
\newblock \emph{ICCV}, 5865--5874.

\bibitem[{Park et~al.(2021{\natexlab{b}})Park, Sinha, Hedman, Barron, Bouaziz, Goldman, Martin-Brualla, and Seitz}]{park2021hypernerf}
Park, K.; Sinha, U.; Hedman, P.; Barron, J.~T.; Bouaziz, S.; Goldman, D.~B.; Martin-Brualla, R.; and Seitz, S.~M. 2021{\natexlab{b}}.
\newblock HyperNeRF: A higher-dimensional representation for topologically varying neural radiance fields.
\newblock \emph{ACM Transactions on Graphics}, 40(6).

\bibitem[{Piala and Clark(2021)}]{piala2021terminerf}
Piala, M.; and Clark, R. 2021.
\newblock Terminerf: Ray termination prediction for efficient neural rendering.
\newblock In \emph{2021 International Conference on 3D Vision (3DV)}, 1106--1114. IEEE.

\bibitem[{Pumarola et~al.(2021)Pumarola, Corona, Pons-Moll, and Moreno-Noguer}]{Pumarola_2021_CVPR}
Pumarola, A.; Corona, E.; Pons-Moll, G.; and Moreno-Noguer, F. 2021.
\newblock D-NeRF: Neural Radiance Fields for Dynamic Scenes.
\newblock In \emph{Proceedings of the IEEE/CVF Conference on Computer Vision and Pattern Recognition (CVPR)}, 10318--10327.

\bibitem[{Shuai et~al.(2022)Shuai, Geng, Fang, Peng, Shen, Zhou, and Bao}]{shuai2022multinb}
Shuai, Q.; Geng, C.; Fang, Q.; Peng, S.; Shen, W.; Zhou, X.; and Bao, H. 2022.
\newblock Novel view synthesis of human interactions from sparse multi-view videos.
\newblock In \emph{SIGGRAPH Conference Proceedings}, 1--10.

\bibitem[{Sitzmann et~al.(2021)Sitzmann, Rezchikov, Freeman, Tenenbaum, and Durand}]{sitzmann2021light}
Sitzmann, V.; Rezchikov, S.; Freeman, B.; Tenenbaum, J.; and Durand, F. 2021.
\newblock Light field networks: Neural scene representations with single-evaluation rendering.
\newblock \emph{Advances in Neural Information Processing Systems}, 34: 19313--19325.

\bibitem[{Suhail et~al.(2022)Suhail, Esteves, Sigal, and Makadia}]{suhail2022light}
Suhail, M.; Esteves, C.; Sigal, L.; and Makadia, A. 2022.
\newblock Light field neural rendering.
\newblock In \emph{Proceedings of the IEEE/CVF Conference on Computer Vision and Pattern Recognition}, 8269--8279.

\bibitem[{Sun et~al.(2022)Sun, Chen, Wang, Li, Averbuch-Elor, Zhou, and Snavely}]{sun2022neural}
Sun, J.; Chen, X.; Wang, Q.; Li, Z.; Averbuch-Elor, H.; Zhou, X.; and Snavely, N. 2022.
\newblock Neural 3D reconstruction in the wild.
\newblock In \emph{ACM SIGGRAPH 2022 Conference Proceedings}, 1--9.

\bibitem[{Tancik et~al.(2022)Tancik, Casser, Yan, Pradhan, Mildenhall, Srinivasan, Barron, and Kretzschmar}]{tancik2022block}
Tancik, M.; Casser, V.; Yan, X.; Pradhan, S.; Mildenhall, B.; Srinivasan, P.~P.; Barron, J.~T.; and Kretzschmar, H. 2022.
\newblock Block-nerf: Scalable large scene neural view synthesis.
\newblock In \emph{Proceedings of the IEEE/CVF Conference on Computer Vision and Pattern Recognition}, 8248--8258.

\bibitem[{Tojo and Umetani(2022)}]{tojo2022posternerf}
Tojo, K.; and Umetani, N. 2022.
\newblock Recolorable posterization of volumetric radiance fields using visibility-weighted palette extraction.
\newblock \emph{Computer Graphics Forum}, 41(4): 149--160.

\bibitem[{Tosic and Berkner(2014)}]{6910019}
Tosic, I.; and Berkner, K. 2014.
\newblock Light field scale-depth space transform for dense depth estimation.
\newblock In \emph{2014 IEEE Conference on Computer Vision and Pattern Recognition Workshops}, 441--448.

\bibitem[{Verbin et~al.(2022)Verbin, Hedman, Mildenhall, Zickler, Barron, and Srinivasan}]{verbin2022ref}
Verbin, D.; Hedman, P.; Mildenhall, B.; Zickler, T.; Barron, J.~T.; and Srinivasan, P.~P. 2022.
\newblock Ref-NeRF: structured view-dependent appearance for neural radiance fields.
\newblock In \emph{2022 IEEE/CVF Conference on Computer Vision and Pattern Recognition (CVPR)}, 5481--5490. IEEE.

\bibitem[{Wang et~al.(2022{\natexlab{a}})Wang, Chandran, Zoss, Bradley, and Gotardo}]{wang2022morf}
Wang, D.; Chandran, P.; Zoss, G.; Bradley, D.; and Gotardo, P. 2022{\natexlab{a}}.
\newblock Morf: Morphable radiance fields for multiview neural head modeling.
\newblock In \emph{ACM SIGGRAPH 2022 Conference Proceedings}, 1--9.

\bibitem[{Wang et~al.(2022{\natexlab{b}})Wang, Ren, Huang, Olszewski, Chai, Fu, and Tulyakov}]{wang2022r2l}
Wang, H.; Ren, J.; Huang, Z.; Olszewski, K.; Chai, M.; Fu, Y.; and Tulyakov, S. 2022{\natexlab{b}}.
\newblock R2l: Distilling neural radiance field to neural light field for efficient novel view synthesis.
\newblock In \emph{Computer Vision--ECCV 2022: 17th European Conference, Tel Aviv, Israel, October 23--27, 2022, Proceedings, Part XXXI}, 612--629. Springer.

\bibitem[{Wang et~al.(2021)Wang, Liu, Liu, Theobalt, Komura, and Wang}]{wang2021neus}
Wang, P.; Liu, L.; Liu, Y.; Theobalt, C.; Komura, T.; and Wang, W. 2021.
\newblock NeuS: Learning neural implicit surfaces by volume rendering for multi-view reconstruction.
\newblock \emph{NeurIPS}, 27171--27183.

\bibitem[{Williams(1978)}]{williams1978casting}
Williams, L. 1978.
\newblock Casting curved shadows on curved surfaces.
\newblock In \emph{Proceedings of the 5th Annual Conference on Computer Graphics and Interactive Techniques}, 270--274.

\bibitem[{Wizadwongsa et~al.(2021)Wizadwongsa, Phongthawee, Yenphraphai, and Suwajanakorn}]{wizadwongsa2021nex}
Wizadwongsa, S.; Phongthawee, P.; Yenphraphai, J.; and Suwajanakorn, S. 2021.
\newblock Nex: Real-time view synthesis with neural basis expansion.
\newblock In \emph{Proceedings of the IEEE/CVF Conference on Computer Vision and Pattern Recognition}, 8534--8543.

\bibitem[{Wu et~al.(2022)Wu, Liu, Chen, Li, Zheng, Cai, and Zheng}]{wu2022object}
Wu, Q.; Liu, X.; Chen, Y.; Li, K.; Zheng, C.; Cai, J.; and Zheng, J. 2022.
\newblock Object-compositional neural implicit surfaces.
\newblock In \emph{Computer Vision--ECCV 2022: 17th European Conference, Tel Aviv, Israel, October 23--27, 2022, Proceedings, Part XXVII}, 197--213. Springer.

\bibitem[{Yang et~al.(2021)Yang, Zhang, Xu, Li, Zhou, Bao, Zhang, and Cui}]{yang2021learning}
Yang, B.; Zhang, Y.; Xu, Y.; Li, Y.; Zhou, H.; Bao, H.; Zhang, G.; and Cui, Z. 2021.
\newblock Learning object-compositional neural radiance field for editable scene rendering.
\newblock In \emph{Proceedings of the IEEE/CVF International Conference on Computer Vision}, 13779--13788.

\bibitem[{Yariv et~al.(2023)Yariv, Hedman, Reiser, Verbin, Srinivasan, Szeliski, Barron, and Mildenhall}]{bakedsdf2023sig}
Yariv, L.; Hedman, P.; Reiser, C.; Verbin, D.; Srinivasan, P.~P.; Szeliski, R.; Barron, J.~T.; and Mildenhall, B. 2023.
\newblock BakedSDF: Meshing Neural SDFs for Real-Time View Synthesis.
\newblock In \emph{ACM SIGGRAPH 2023 Conference Proceedings}, SIGGRAPH '23. New York, NY, USA: Association for Computing Machinery.
\newblock ISBN 9798400701597.

\bibitem[{Yu et~al.(2021)Yu, Li, Tancik, Li, Ng, and Kanazawa}]{yu2021plenoctrees}
Yu, A.; Li, R.; Tancik, M.; Li, H.; Ng, R.; and Kanazawa, A. 2021.
\newblock Plenoctrees for real-time rendering of neural radiance fields.
\newblock In \emph{Proceedings of the IEEE/CVF International Conference on Computer Vision}, 5752--5761.

\bibitem[{Zhang et~al.(2021)Zhang, Srinivasan, Deng, Debevec, Freeman, and Barron}]{zhang2021nerfactor}
Zhang, X.; Srinivasan, P.~P.; Deng, B.; Debevec, P.; Freeman, W.~T.; and Barron, J.~T. 2021.
\newblock Nerfactor: Neural factorization of shape and reflectance under an unknown illumination.
\newblock \emph{ACM Transactions on Graphics (TOG)}, 40(6): 1--18.

\bibitem[{Zhang et~al.(2022)Zhang, Sun, He, Fu, Jia, and Zhou}]{zhang2022modeling}
Zhang, Y.; Sun, J.; He, X.; Fu, H.; Jia, R.; and Zhou, X. 2022.
\newblock Modeling indirect illumination for inverse rendering.
\newblock In \emph{Proceedings of the IEEE/CVF Conference on Computer Vision and Pattern Recognition}, 18643--18652.

\end{thebibliography}

\appendixpage

\section{Intersection Network}

\begin{figure}[htbp]
    \centering
    \includegraphics[width=\linewidth]{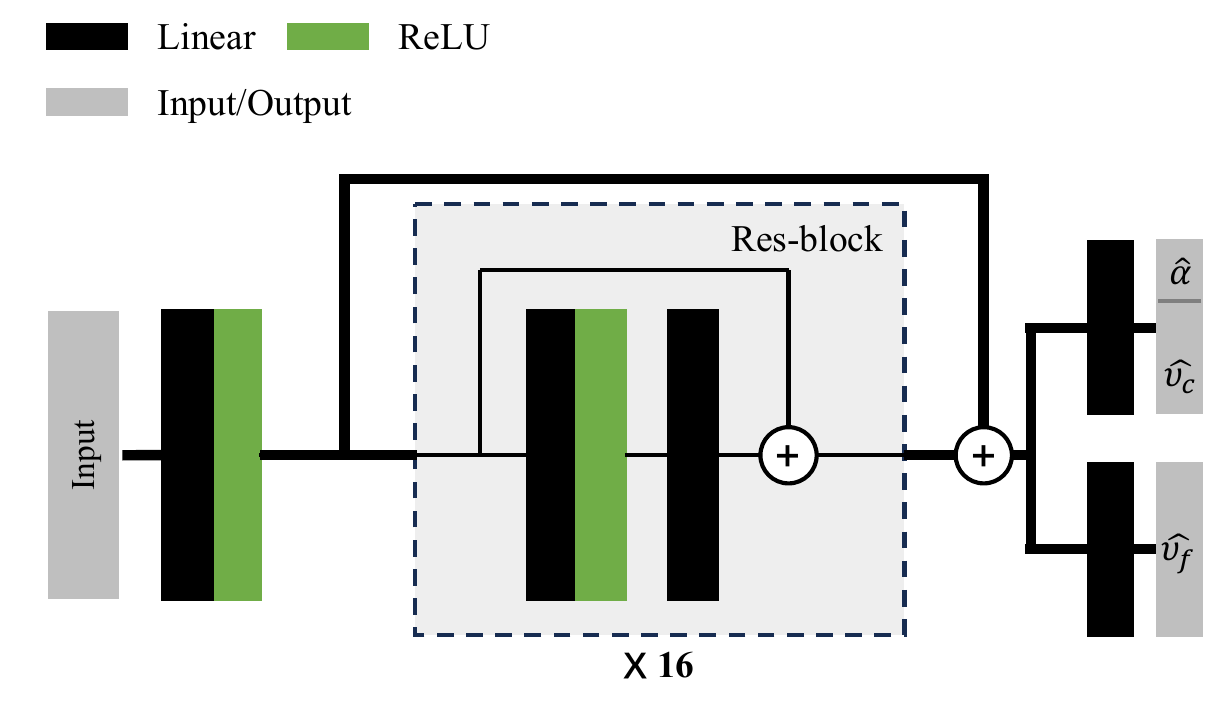}
    \caption{Architecture of Intersection Network.}
    \label{fig:nedf_arch}
\end{figure}

\paragraph{Architecture.} The Intersection Network is built using an MLP (Fig. \ref{fig:nedf_arch}). The MLP's head is a single linear layer that converts the input dimension $D_i$ to the feature dimension $D_f$. The MLP's body is made up of 16 residual blocks, each of which is made up of two fully connected layers followed by ReLU activation. The MLP's tail has two branches, each with a single linear to fit output dimension. One branch produces $\hat{\alpha}$ and $\hat{\upsilon}_c$, while the other produces $\hat{\upsilon}_f$ (Equation 3 in the main paper).

\paragraph{The choice for hyper-parameters.} In our experiment, $D_i$ is 1,008 (16 points for a ray, each with 10-level sinusoidal position encoding, as in NeRF). We set $D_f$ to 256 and use 16 residual blocks with skip-connection, following the same idea as Wang et al. \shortcite{wang2022r2l}, where residual blocks with skip-connection play an important role in deep networks. More residual blocks result in a slight improvement in quality but a significant decrease in speed, and 16 blocks is a good choice for trade-off in most cases.

\paragraph{Training details.} We employ Adam as the optimizer and set the learning rate to 5e-4. We use the pre-trained NeRF models to generate 500 random views (more random views, less artifacts) for supervision for each NeDF model of a single object, and train the intersection network over 60W iterations until convergence. The batchsize of training rays is set to 4,096 during each iteration.

\section{Algorithm Pseudo Code}
We detailed algorithm \ref{algo:one} and \ref{algo:two} in pseudo-code corresponding to the three main steps in our pipeline.

\begin{small}
\begin{algorithm}[htbp]
\SetAlgoLined
    \caption{Fast NeRF Composition}
    \label{algo:one}
    \KwIn{
    $\mathcal{M}$, all models in the view frustum. Each model has a unique id\;
    }
    \KwOut{color buffer $I_{rgb}$ of compositing results}
    \BlankLine
    \textbf{NeDF Generation Step (STEP 1)}; Output: depth buffer $I_{depth}$, id buffer $I_{id}$\;
    \emph{clear depth buffer with $\infty$, reset id buffer}\;
    \ForEach{model $m$ of $\mathcal{M}$}{
        use $\mathcal{F}_{\Phi}$ of $m$\;
        \ForEach{ray cast through pixel $i$}{
            render depth $d$ and mask $\alpha$ by $\mathcal{F}_{\Phi}$\;
            \If{$d < I_{depth}^i$ and $\alpha=1$ }{
                write to $I_{depth}^i$ with $d$\;
                write to $I_{id}^i$ with $m.id$\;
            }
        } 
    }
    \BlankLine
    \textbf{Deferred Shading Step (STEP 2)}\;
    \emph{clear color buffer with \textit{clear-color value}}\;
    \ForEach{model $m$ of $\mathcal{M}$}{
        use $\mathcal{F}_{\Theta}$ of $m$\;
        \ForEach{pixel $i$ where $I_{id}^i==m.id$}{
            project pixel $i$ to a single 3D point $\mathbf{x}_i$ using $I_{depth}^i$\;
            render color at $\mathbf{x}_i$ by $F_{\Theta}$ and write to $I_{rgb}^i$\;
        } 
    }
    \BlankLine
\end{algorithm}
\end{small}

\begin{algorithm}[htbp]
\SetAlgoLined
    \caption{Fast Dynamic Shadow Casting}
    \label{algo:two}
    \KwIn{
    $\mathcal{M}$, all Models in view frustrum. Each model has a unique id. $L$, position of a point light source\;
    }
    \KwOut{color buffer $I_{rgb}$ with shadows}
    \BlankLine
    \textbf{Shadow Step (STEP 3)}: generate shadow by shadow rays\;
    \ForEach{pixel $i$ with a valid $I_{id}^i$}{
        project pixel $i$ to a single 3D point $\mathbf{x}_i$ using $I_{depth}^i$\;
        cast shadow ray from $L$ to $\mathbf{x}_i$\;
        depth $D$ in shadow maps can be obtained like in STEP 1\;
        \If{$D+\epsilon<|L-\mathbf{x}_i|$}{
            multiply $I_{rgb}^i$ with an intensity value $\beta(0<\beta<1)$\;
        }
    } 
    \BlankLine
\end{algorithm}

\section{A brief Comparison Result}
\begin{figure*}[htbp]
  \includegraphics[width=\linewidth]{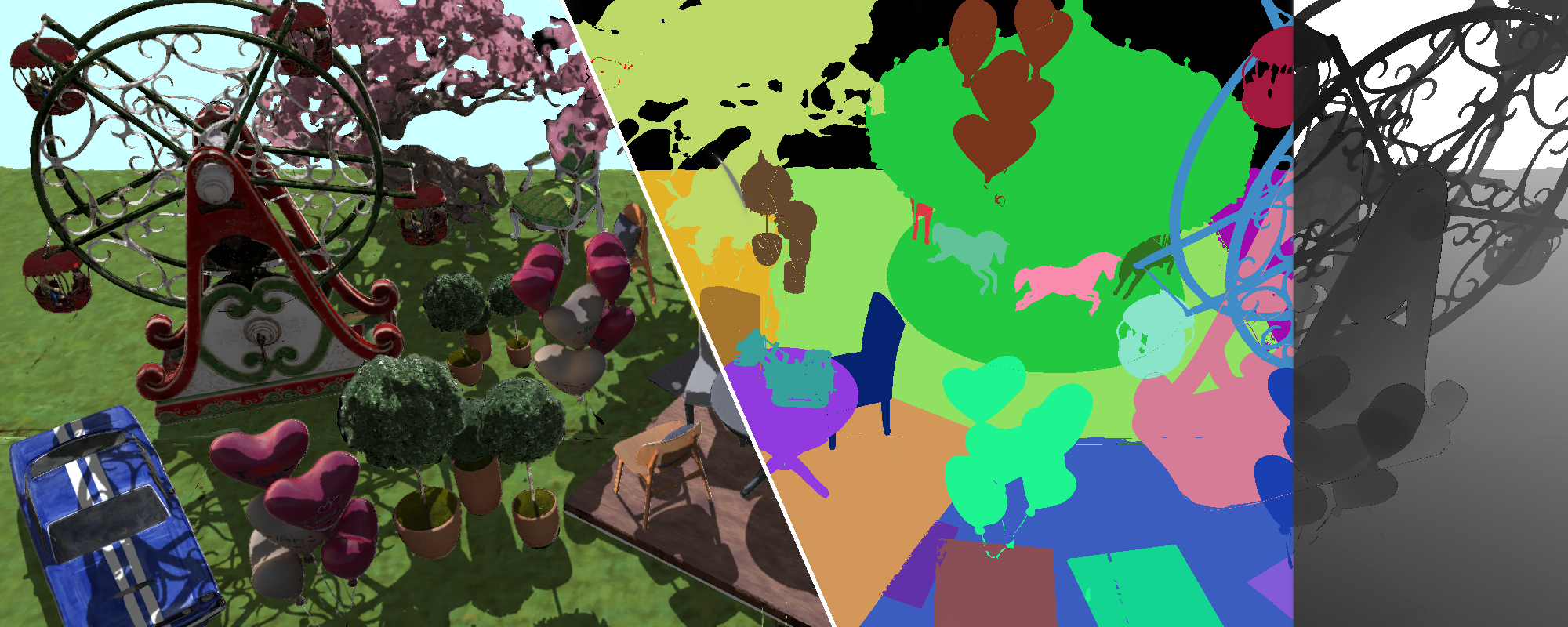}
  \caption{
  The above frame (with shadow) with a resolution of 2,000x800 was rendered in 4.48 seconds using our Pytorch implementation, whereas vanilla NeRF takes over 130 seconds and the hybird representation based NeRF method NSVF \cite{liu2020neural} takes over 10 seconds (without the presence of a shadow). The above scene is dynamic, e.g., a carousel and a ferris wheel that both rotate slowly (see the demo video in the supplemental materials for more details).
  }
  \label{fig:teaser}
\end{figure*}

We force a brief comparison with NSVF \cite{liu2020neural} in caption of Fig. \ref{fig:teaser} (the most complex and dynamic scene from our dataset). Because NSVF and ours are two totally different system, it's hard to set a simultaneous event and we only made a comparison in terms of speed.

\end{document}